\documentclass{article}

\usepackage{iclr2027_conference,times}

\usepackage{amsmath,amsfonts,bm}

\def\eqref#1{equation~\ref{#1}}

\def\1{\bm{1}}

\DeclareMathAlphabet{\mathsfit}{\encodingdefault}{\sfdefault}{m}{sl}
\SetMathAlphabet{\mathsfit}{bold}{\encodingdefault}{\sfdefault}{bx}{n}

\newcommand{\E}{\mathbb{E}}

\newcommand{\KL}{D_{\mathrm{KL}}}
\newcommand{\Var}{\mathrm{Var}}

\newcommand{\Cov}{\mathrm{Cov}}

\renewcommand{\eqref}[1]{\textup{(\ref{#1})}}

\usepackage{hyperref}
\usepackage{url}
\usepackage{todonotes}
\usepackage{bm}
\usepackage{algorithm}
\usepackage{booktabs}
\usepackage{graphicx}
\usepackage{placeins}
\usepackage{float}
\usepackage{pdflscape}
\usepackage{algpseudocode}
\usepackage{amsthm}
\usepackage{xcolor}
\usepackage{colortbl}
\usepackage{capt-of}
\usepackage{arydshln}
\definecolor{taskband}{RGB}{241,245,248}
\definecolor{brightorange}{HTML}{FF8C00}
\definecolor{softorange}{RGB}{230,145,70}
\definecolor{softpurple}{HTML}{9B7EDB}
\hypersetup{citebordercolor=softorange,linkbordercolor=softpurple}

\newcommand{\Normal}{\mathcal{N}}

\newcommand{\mat}[1]{\bm{#1}}

\DeclareMathOperator{\sg}{sg}
\newtheorem{proposition}{Proposition}
\newcommand{\vct}[1]{\bm{#1}}
\newcommand{\norm}[1]{\left\|#1\right\|}
\title{FASTER: Fast Adjoint Stochastic Transport for Endpoint Refinement in Reward-Guided Image Editing}

\iclrfinalcopy
\author{
Yimiao Zhou\textsuperscript{1}\thanks{Equal contribution.
\qquad \textsuperscript{\textdagger}Corresponding author.}
\hspace{1.5em}
Zejia Zhong\textsuperscript{1}\footnotemark[1]
\hspace{1.5em}
Jingya Wang\textsuperscript{1}
\hspace{1.5em}
Ye Shi\textsuperscript{1\textdagger} \\
\textsuperscript{1}ShanghaiTech University \\
Shanghai, China \\
\texttt{\{zhouym2025,zhongzj2025,wangjingya,shiye\}@shanghaitech.edu.cn}
}

\begin{document}
\raggedbottom
\maketitle

\begin{abstract}
Reward-guided image editing at test time seeks to improve a specified reward while preserving source content and visual plausibility. Many existing approaches optimize candidates through pretrained generation processes, making repeated adjustment depend on costly large-model execution and, in some cases, backbone backpropagation. We develop a theoretical framework that jointly accounts for reward, source preservation, and pretrained-prior preferences, allowing the desired output distribution to be specified separately from the dynamics used to realize it. Based on this framework, we introduce FASTER, which trains a small network for each source and objective to perform inexpensive editing, while pretrained and reward models provide feedback on candidate outputs. By reusing each candidate and its feedback across multiple small-network updates, FASTER reduces repeated sampling and supervision queries without placing the pretrained generative backbone inside the inner optimization loop. On SD3, FASTER leads all four target metrics and several validation metrics among the evaluated methods. Compared with the evaluated baseline that optimizes controls along pretrained generation trajectories, FASTER achieves editing-time speedups of up to \({6.91\times}\) on Stable Diffusion 3 and \({24.14\times}\) on Stable Diffusion 1.5.
\end{abstract}

\section{Introduction}
\label{sec:introduction}

\suppressfloats[t]
\begin{figure}[t]
\centering
\includegraphics[width=0.92\linewidth,trim=2 2 2 2,clip]{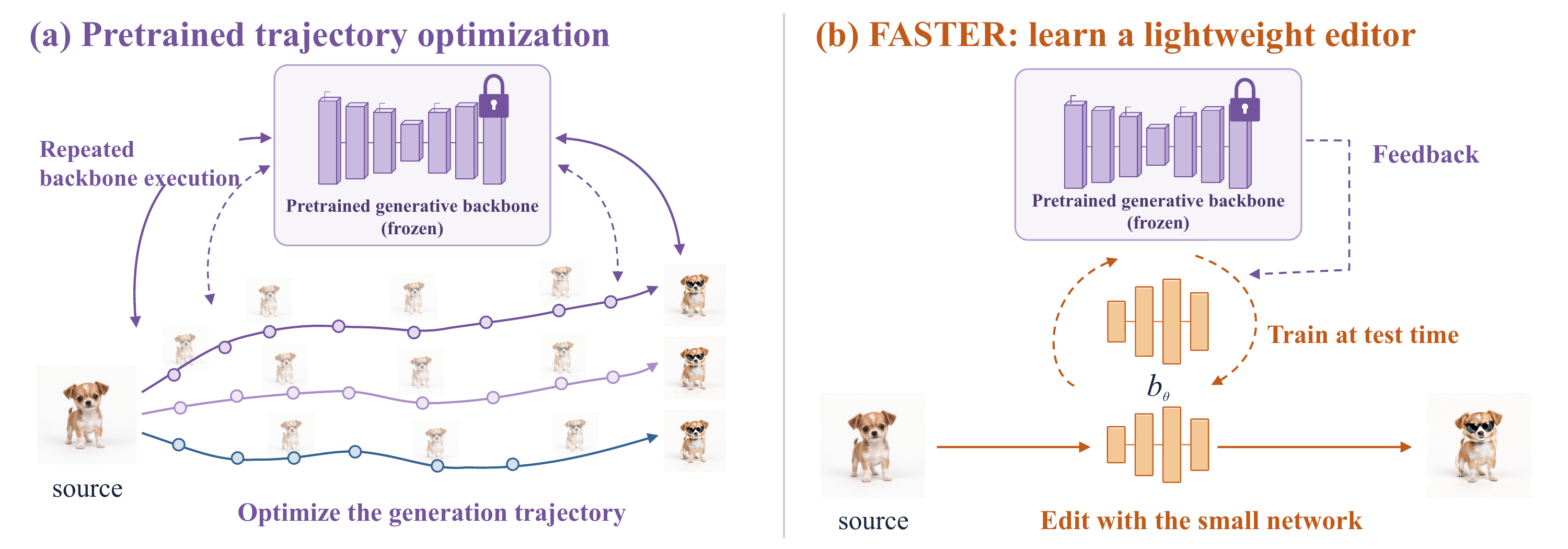}
\vspace{-4mm}
\caption{\textbf{Decoupling pretrained knowledge from editing dynamics.} (a) Trajectory optimization repeatedly executes a frozen generative backbone. (b) FASTER learns a small per-image editor at test time, reusing frozen-model feedback across updates; the learned network then performs the edit.}
\label{fig:motivation}
\end{figure}

Pretrained diffusion and flow models provide rich visual priors for image generation~\citep{ho2020ddpm,song2020score,rombach2022ldm,lipman2023flow,liu2023rectifiedflow,esser2024sd3}. However, their conditioning interfaces do not fully express every downstream objective: human preferences, reference styles, and task-specific semantic requirements may instead be specified through rewards available at test time. Reward-guided inference adapts generated outputs to such objectives without retraining the pretrained model~\citep{bansal2023universal,yu2023freedom,ye2024tfg,tang2025dno,eyring2024reno,susladkar2026rewardflow}. We focus on \emph{reward-guided image editing at test time}: given a source image and a reward objective, improve the objective while retaining content that need not change and maintaining visual plausibility~\citep{chang2026itoc}. Unlike unconstrained reward maximization, this task requires meaningful, source-preserving changes; a higher reward alone does not establish a successful edit.

Existing approaches typically inject reward guidance into pretrained generation or optimize candidates through a generation mapping or trajectory. Guidance methods use external gradients to adjust intermediate states, with refinements to the guidance geometry and update rules~\citep{chung2023dps,bansal2023universal,yu2023freedom,he2024mpgd,yang2024dsg,ye2024tfg,susladkar2026rewardflow}. Optimization-based approaches instead adjust input noise, editable variables, or trajectory controls to improve the generated output~\citep{wallace2023doodl,benhamu2024dflow,tang2025dno,eyring2024reno,chang2026itoc}. Other advances improve efficiency through reward optimization of one-step models, flow-based editing, or efficient reward guidance~\citep{eyring2024reno,patel2025flowchef,huang2026fmrg}. Despite their different computational costs, many of these approaches couple the pretrained model's role as a source of visual knowledge with its role in executing candidate updates. Repeated adjustment invokes the generator again, and some approaches additionally backpropagate through its backbone; candidate evaluation introduces further decoding and reward-computation costs. Can pretrained knowledge continue to constrain desirable edits while a separate, lightweight process performs the actual adjustment?

Figure~\ref{fig:motivation} contrasts optimization along pretrained generation trajectories with FASTER's lightweight editing process: the former repeatedly executes the large generative backbone, whereas the latter uses a small network to perform editing with feedback from frozen models. We establish a theoretical basis for this separation by jointly analyzing reward improvement, source preservation, and pretrained-prior plausibility, specifying a target output distribution and constructing an independent editing process that realizes it under ideal conditions. Each task fixes a source and an objective, so the learned editor need only adapt the current instance rather than acquire general-purpose generation capabilities.

Based on this framework, we introduce \textbf{FASTER}: a small network trained at test time performs inexpensive editing updates, while pretrained and reward models provide feedback at candidate outputs. Reusing each candidate and its feedback across multiple network updates reduces repeated sampling and costly supervision queries. To support explicit semantic changes that increasing an image--text score may fail to produce, we further introduce an editing term that obtains semantic directions from the pretrained model's conditional predictions, without backpropagation through its backbone. This term enters the same feedback interface, allowing the lightweight editor to use pretrained semantic knowledge while retaining its inexpensive update process. On Stable Diffusion 3~\citep{esser2024sd3}, FASTER leads the target metric in all four tasks and several validation metrics among the evaluated methods. Relative to the evaluated baseline that optimizes controls along pretrained generation trajectories, it achieves editing-time speedups of up to $6.91\times$ on Stable Diffusion~3~\citep{esser2024sd3} and $24.14\times$ on Stable Diffusion~1.5~\citep{rombach2022ldm}.

Our contributions are threefold:
\begin{itemize}
\setlength{\itemsep}{2pt}
\item \textbf{Decoupling the editing target from its dynamics.} We combine reward, source preservation, and pretrained-prior preferences into a target distribution, and construct separate lightweight dynamics that realize it under proper conditions theoretically.
\item \textbf{FASTER: inexpensive editing with a small network.} A per-image small network executes inexpensive edits and reuses candidate feedback to reduce repeated computation. A semantic editing term supplies directions from conditional model predictions through the same interface, without generative-backbone backpropagation.
\item \textbf{Editing quality and efficiency across tasks.} FASTER leads all four target metrics and several validation metrics in the SD3 comparison, with up to $\boldsymbol{6.91\times}$ and $\boldsymbol{24.14\times}$ speedups on SD3 and SD1.5, respectively. A $768\times768$ SD3 ImageReward edit takes approximately $7.7$ seconds on a single consumer GPU.
\end{itemize}

\section{Related Work}
\label{sec:related}

\paragraph{Inference-time guidance and alignment.}
Training-free guidance steers pretrained diffusion and flow models toward external objectives without updating model weights.
DPS, Universal Guidance, and FreeDoM construct guidance from denoised predictions or external energies~\citep{chung2023dps,bansal2023universal,yu2023freedom}, while MPGD, DSG, and TFG further improve the geometry and design of guidance~\citep{he2024mpgd,yang2024dsg,ye2024tfg}.
Alternatively, DOODL and D-Flow optimize latent variables through the pretrained generative mapping~\citep{wallace2023doodl,benhamu2024dflow}, with DNO and ReNO improving robustness or efficiency of direct optimization~\citep{tang2025dno,eyring2024reno}.
Distributional methods such as DAS and Feynman--Kac steering instead reshape inference through reweighting and resampling~\citep{kim2025das,singhal2025fk}.
These approaches provide flexible test-time alignment, but generally remain coupled to pretrained generation during optimization.

\paragraph{Training-free and reward-guided image editing.}
Training-free image editing must also preserve source correspondence.
Existing methods use inversion and attention control~\citep{hertz2023p2p,mokady2023nti}, controlled corruption~\citep{meng2022sdedit}, and feature reuse~\citep{tumanyan2023pnp,cao2023masactrl}.
Other approaches rely on score differences~\citep{hertz2023dds} or directly model source-to-target editing dynamics~\citep{xu2024infedit,kulikov2025flowedit}.
For reward-guided editing, ITOC optimizes controls along the pretrained trajectory~\citep{chang2026itoc}.
FlowChef exploits flow geometry~\citep{patel2025flowchef}, whereas RewardFlow combines multiple differentiable rewards~\citep{susladkar2026rewardflow}.
FMRG further improves guidance efficiency through flow-map lookahead~\citep{huang2026fmrg}. FASTER instead specifies a source-conditioned endpoint target and learns a separate lightweight transport to realize it.

\paragraph{Generative priors and stochastic-control optimization.}
Frozen models provide priors for inverse imaging~\citep{feng2023priors,feng2024surrogate,mardani2024reddiff} and score-distillation-based 3D optimization~\citep{poole2023dreamfusion,wang2023prolificdreamer}.
UniDB formulates diffusion bridges as SOC with a tunable terminal penalty~\citep{zhu2025unidb}, while UniDB++ accelerates sampling within this framework through a training-free solver~\citep{pan2025unidbplusplus}.
For SOC solvers, Adjoint Matching learns controls by regression~\citep{domingoenrich2025adjoint}, Adjoint Sampling reuses terminal information through reference bridges~\citep{havens2025adjointsampling}, and Efficient Adjoint Matching simplifies the reference dynamics and terminal cost~\citep{shin2026eam}.
FASTER uses frozen-model endpoint supervision to learn a lightweight per-image editor.
\section{Preliminaries}
\label{sec:prelim}

We edit in latent space: $\vct{s}$ and $\vct{y}$ denote source and output latents, $\mathcal D$ is a frozen decoder, and $r(\vct{y})=R(\mathcal D(\vct{y}))$ is a task reward. We write $p_{\mathrm{pre}}$ for the pretrained latent prior. The control time $\lambda\in[0,1]$ below is distinct from the pretrained model's noise time $t$ used for semantic queries in Sec.~\ref{sec:method}.

\subsection{Reference Dynamics and Stochastic Control}
\label{sec:prelim_soc}

A reference process describes state evolution before applying a task-dependent control. From a fixed initial state $\vct{x}_{\mathrm{init}}$, its path law $Q$ and a controlled path law $P^{\vct{b}}$ are defined by
\begin{subequations}
\label{eq:general-control}
\begin{align}
Q:&&d\vct{x}_\lambda&=\vct{a}_0(\vct{x}_\lambda,\lambda)\,d\lambda+\sigma_\lambda\,d\vct{w}_\lambda,\\
P^{\vct{b}}:&&d\vct{x}_\lambda&=[\vct{a}_0(\vct{x}_\lambda,\lambda)+\vct{b}(\vct{x}_\lambda,\lambda)]\,d\lambda+\sigma_\lambda\,d\vct{w}_\lambda,
\qquad \vct{x}_0=\vct{x}_{\mathrm{init}}.
\end{align}
\end{subequations}
Here $\vct{a}_0$ is the reference drift, $\vct{b}$ the control drift, and $\sigma_\lambda>0$ the shared diffusion scale. In pretrained-trajectory control, $\vct{a}_0$ contains the frozen generator's prediction, so each state update still invokes the large model even when the control branch is small. ITOC illustrates this computational dependence~\citep{chang2026itoc}; its fixed-noise trajectory optimization is distinct from the distributional formulation below.

For a terminal cost $g$, stochastic optimal control minimizes
\begin{equation}
J(\vct{b})=\E_{P^{\vct{b}}}\!\left[\int_0^1\frac{\|\vct{b}(\vct{x}_\lambda,\lambda)\|^2}{2\sigma_\lambda^2}\,d\lambda+g(\vct{x}_1)\right]
=\KL(P^{\vct{b}}\Vert Q)+\E_{P^{\vct{b}}}[g(\vct{x}_1)].
\label{eq:soc-general}
\end{equation}
The equality requires the usual Girsanov and integrability conditions~\citep{domingoenrich2025adjoint,havens2025adjointsampling}. The reference process regularizes paths, while $g$ specifies the preference over endpoints. These are separate design choices: the pretrained prior $p_{\mathrm{pre}}$ can inform the endpoint preference without dictating the reference dynamics $Q$.

\subsection{Adjoint Matching}
\label{sec:prelim_am}

Adjoint Matching (AM) trains a control by regressing it toward an adjoint-derived target~\citep{domingoenrich2025adjoint}. For a sampled trajectory held fixed during regression, its lean adjoint in the terminal-cost setting obeys
\begin{equation}
\frac{d\widetilde{\vct{a}}_\lambda}{d\lambda}
=-\bigl[\nabla_{\vct{x}}\vct{a}_0(\vct{x}_\lambda,\lambda)\bigr]^\top\widetilde{\vct{a}}_\lambda,
\qquad \widetilde{\vct{a}}_1=\nabla g(\vct{x}_1),
\label{eq:prelim-lean-adjoint}
\end{equation}
and supplies the physical-drift regression target $-\sigma_\lambda^2\widetilde{\vct{a}}_\lambda$. AM is a solver for the control problem, not another definition of its objective. Although its lean adjoint avoids differentiating the learned control, Eq.~\eqref{eq:prelim-lean-adjoint} retains derivatives of the reference drift. A pretrained reference can therefore incur large-model work both when sampling paths and when propagating supervision. This motivates the choice of reference dynamics in Sec.~\ref{sec:method}.

\section{FASTER}
\label{sec:method}

\begin{figure}[t]
    \centering
    \includegraphics[
        width=\linewidth,
        trim=26 7 26 78,
        clip
    ]{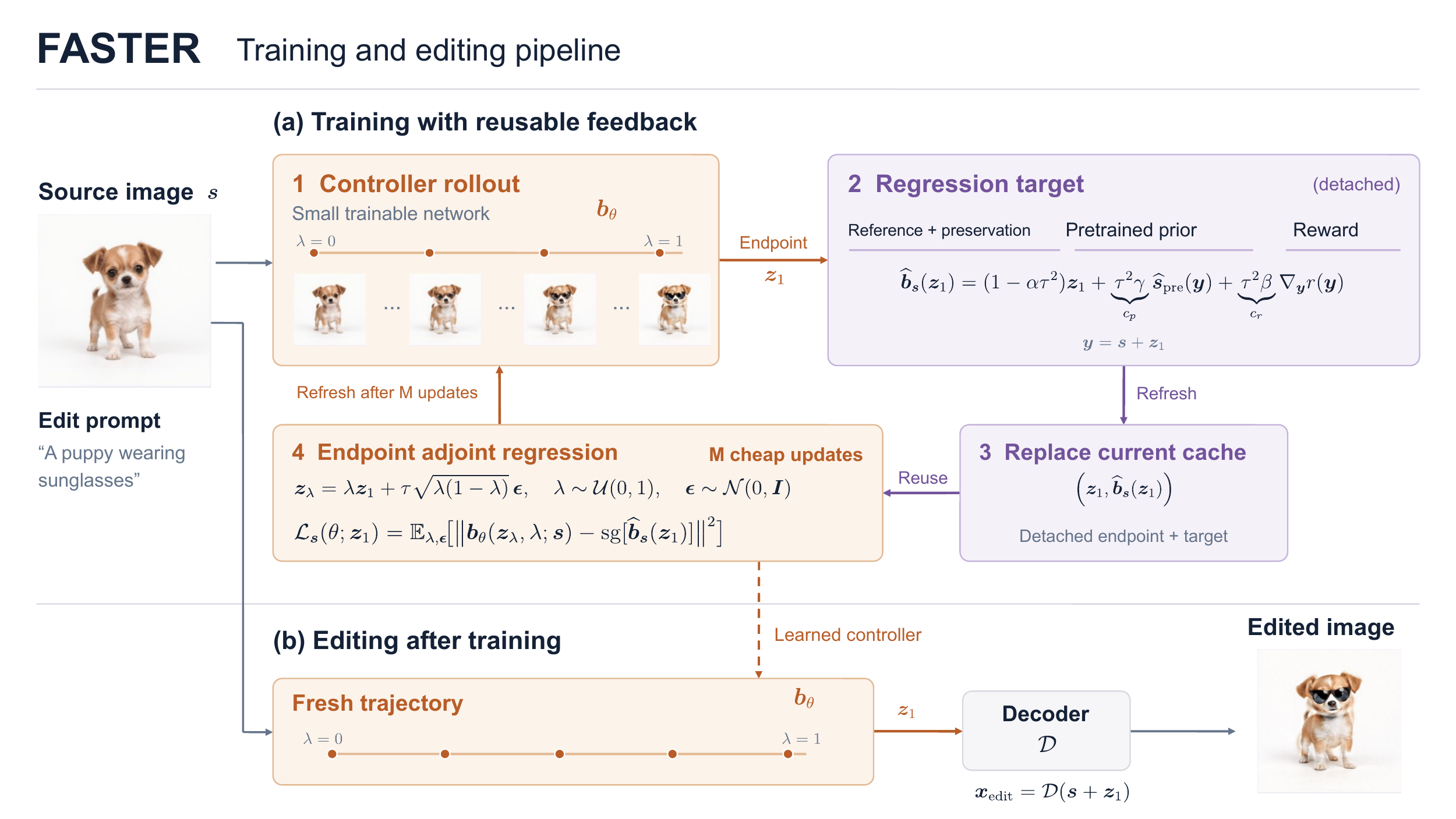}
    \vspace{-10pt}
    \caption{\textbf{FASTER training and editing pipeline.} (a) Each of $K$ refreshes caches an endpoint and its detached preservation, prior, and reward target for $M$ controller updates. (b) After training, a fresh controller rollout is decoded into the edited image.}
    \label{fig:pipeline}
\end{figure}

Pretrained-trajectory control repeatedly executes the large generator as the edit changes (Sec.~\ref{sec:prelim}). FASTER instead specifies a per-image endpoint target and learns a small controller to reach it, reusing endpoint feedback across controller updates (Fig.~\ref{fig:pipeline}).

\subsection{From a Per-Image Editing Objective to Lightweight Control}
\label{sec:method-target}

For a fixed source latent $\vct{s}$, a successful edit should improve the task reward $r(\vct{y})$, avoid unnecessary changes to the source, and favor plausible outputs under the pretrained prior $p_{\mathrm{pre}}$. Let $D(\vct{s},\vct{y})$ measure source change and $\rho$ be an output density for this editing instance. We minimize the following functional over output densities:
\begin{equation}
F_{\vct{s}}(\rho)=\KL(\rho\Vert p_{\mathrm{pre}})
+\alpha\E_\rho[D(\vct{s},Y)]-\beta\E_\rho[r(Y)],
\qquad \alpha\geq0,\quad\beta>0.
\label{eq:editing-objective}
\end{equation}
The KL term favors plausible outputs, while the preservation cost anchors them to $\vct{s}$. The minimizer is the target below with $\gamma=1$; an independent weight $\gamma\geq0$ controls the prior contribution (Appendix~\ref{app:conditional}):
\begin{equation}
\pi_{\vct{s}}(\vct{y})=Z_{\vct{s}}^{-1}p_{\mathrm{pre}}(\vct{y})^\gamma
\exp[-\alpha D(\vct{s},\vct{y})+\beta r(\vct{y})].
\label{eq:target}
\end{equation}
We assume $0<Z_{\vct{s}}<\infty$ and use $D(\vct{s},\vct{y})=\|\vct{y}-\vct{s}\|^2/2$. This target specifies preferred outputs without prescribing their generation paths.

For a fixed source and request, the source supplies content and frozen models supply feedback. A small per-image controller therefore need only learn the current edit, rather than the full image distribution.

Let $\vct{z}_\lambda$ be the residual from $\vct{s}$ and $\tau>0$ a noise scale. We choose a zero-drift Brownian reference $Q_\tau$. Its endpoint density and endpoint-conditioned bridge are analytic:
\begin{subequations}
\label{eq:brownian-reference}
\begin{align}
d\vct{z}_\lambda&=\tau\,d\vct{w}_\lambda,\quad \vct{z}_0=0,
\quad q_\tau(\vct{z})=\Normal(\vct{z};0,\tau^2\mat I),\\
\vct{z}_\lambda\mid\vct{z}_1&\sim\Normal(\lambda\vct{z}_1,\tau^2\lambda(1-\lambda)\mat I),\quad 0<\lambda<1.
\end{align}
\end{subequations}
Adding the per-image controller gives the editing dynamics
\begin{equation}
d\vct{z}_\lambda=\vct{b}_\theta(\vct{z}_\lambda,\lambda;\vct{s})\,d\lambda
+\tau\,d\vct{w}_\lambda,\qquad \vct{y}_\lambda=\vct{s}+\vct{z}_\lambda.
\label{eq:faster-sde}
\end{equation}
We implement the controller with a small network fitted separately for each source and objective:
\begin{equation}
\vct{b}_{\theta}(\vct{z},\lambda;\vct{s})
=\vct{m}+f_\psi(\vct{z},\lambda,\vct{s}),\qquad\theta=(\vct{m},\psi).
\label{eq:parameterization}
\end{equation}
The latent-shaped offset $\vct{m}$ is independent of state and time. The three-layer convolutional network $f_\psi$ takes the residual, source latent, and broadcast time as input and predicts a correction. Each editing step evaluates only this small network; only $\theta$ is trained, with frozen pretrained models providing endpoint feedback. Architecture and initialization details are in Appendix~\ref{app:implementation}.

The reference endpoint $q_\tau$ is Gaussian, however, so it need not equal the desired residual density $\bar\pi_{\vct{s}}(\vct{z})=\pi_{\vct{s}}(\vct{s}+\vct{z})$. We compensate with the terminal cost
\begin{equation}
g_{\vct{s}}(\vct{z})=\log q_\tau(\vct{z})-\log\bar\pi_{\vct{s}}(\vct{z}).
\label{eq:faster-soc}
\end{equation}
Since $q_\tau e^{-g_{\vct{s}}}=\bar\pi_{\vct{s}}$, the unrestricted optimum of Eq.~\eqref{eq:soc-general} has $\vct{y}_1\sim\pi_{\vct{s}}$ under the conditions in Appendix~\ref{app:theory}. The terminal cost cancels the reference endpoint preference and imposes the editing target; a finite controller with estimated feedback approximates this ideal law.

\subsection{Learning from Reusable Endpoint Supervision}
\label{sec:method-solver}

Endpoint feedback requires decoding, reward gradients, and pretrained-model queries. We reuse each evaluated endpoint for multiple small-network updates, reducing repeated rollouts and feedback computation.

The optimal path law $P_{\vct{s}}^\star$ for Eq.~\eqref{eq:faster-soc} has density ratio $dP_{\vct{s}}^\star/dQ_\tau=\bar\pi_{\vct{s}}(\vct{z}_1)/q_\tau(\vct{z}_1)$, which depends only on the endpoint. Thus its conditional paths remain the reference bridges in Eq.~\eqref{eq:brownian-reference} (Appendix~\ref{app:theory}). This structure is useful before convergence as well. For a current path law $P$ with endpoint density $p_1$, the path-space objective decomposes as
\begin{equation}
J_{\vct{s}}(P)=\KL(p_1\Vert\bar\pi_{\vct{s}})
+\E_{\vct{z}_1\sim p_1}\!\left[\KL\!\left(P(\cdot\mid\vct{z}_1)\Vert Q_\tau(\cdot\mid\vct{z}_1)\right)\right].
\label{eq:bridge-kl-decomposition}
\end{equation}
Replacing the conditional paths by reference bridges preserves the endpoint distribution and removes the second term. This motivates pairing current endpoints with resampled bridge states. We adopt Reciprocal Adjoint Matching from Adjoint Sampling~\citep{havens2025adjointsampling}, whose reciprocal-projection analysis supports this construction.

The Brownian reference in Eq.~\eqref{eq:brownian-reference} has $\vct{a}_0=0$, hence $\nabla_{\vct{x}}\vct{a}_0=0$; the lean-adjoint equation~\eqref{eq:prelim-lean-adjoint} therefore gives $d\widetilde{\vct{a}}_\lambda/d\lambda=0$ and $\widetilde{\vct{a}}_\lambda=\nabla g_{\vct{s}}(\vct{z}_1)$. Its physical-drift label is therefore $-\tau^2\nabla g_{\vct{s}}(\vct{z}_1)$, eliminating backward adjoint propagation. For quadratic preservation, the practical label at $\vct{y}=\vct{s}+\vct{z}_1$ is
\begin{equation}
\widehat{\vct{b}}_{\vct{s}}(\vct{z}_1)
=(1-\alpha\tau^2)\vct{z}_1
+\underbrace{\tau^2\gamma}_{c_p}\widehat{\vct{s}}_{\mathrm{pre}}(\vct{y})
+\underbrace{\tau^2\beta}_{c_r}\nabla_{\vct{y}}r(\vct{y}).
\label{eq:quadratic-label}
\end{equation}
The first term combines reference correction and preservation; the remaining terms provide estimated prior and reward gradients. Here $\widehat{\vct{s}}_{\mathrm{pre}}$ is a prior-guidance estimate obtained from noisy model queries; Appendix~\ref{app:oracle} describes its surrogate-prior interpretation. Reward gradients pass through the frozen decoder and reward model at this endpoint. We evaluate the label once, then store it with the detached endpoint.

For the stored pair, draw $\lambda\sim\mathcal U(0,1)$ and independent $\vct{\epsilon}\sim\Normal(0,\mat I)$. Sample the reference bridge as $\vct{z}_\lambda=\lambda\vct{z}_1+\tau\sqrt{\lambda(1-\lambda)}\,\vct{\epsilon}$ and minimize
\begin{equation}
\mathcal L_{\vct{s}}(\theta;\vct{z}_1)=\E_{\lambda,\vct{\epsilon}}\!\left[\left\|\vct{b}_\theta(\vct{z}_\lambda,\lambda;\vct{s})-\sg[\widehat{\vct{b}}_{\vct{s}}(\vct{z}_1)]\right\|^2\right].
\label{eq:ram-loss}
\end{equation}
The label and sampled states are detached. Resampling bridge states supplies new training inputs without new rollouts or feedback queries. Appendix~\ref{app:solver} establishes population consistency under exact target endpoints and labels, and discusses finite-update limitations.

Each of $K$ refreshes replaces the cached endpoint--label pair and performs $M$ controller updates. After fitting, a fresh controller sample is decoded as the output (Algorithm~\ref{alg:faster}). Appendix~\ref{app:feedback-reuse} examines the effect of varying $K$ and $M$.

\begin{algorithm}[t]
\caption{FASTER: per-image endpoint refinement}
\label{alg:faster}
\begin{algorithmic}[1]
\Require Source latent $\vct{s}$; reward $r$; $\alpha,\tau,c_p,c_r$; refreshes $K$; updates $M$.
\State Initialize $\vct{m}$ and the final layer of $f_\psi$ to zero in Eq.~\eqref{eq:parameterization}.
\For{$k=1,\ldots,K$}
    \State Roll out an endpoint $\vct{z}_1$ using Eq.~\eqref{eq:faster-sde}.
    \State Evaluate Eq.~\eqref{eq:quadratic-label}; replace the cache with detached $(\vct{z}_1,\widehat{\vct{b}}_{\vct{s}}(\vct{z}_1))$.
    \For{$j=1,\ldots,M$}
        \State Draw a minibatch of independent $(\lambda,\vct{\epsilon})$ pairs.
        \State Form detached $\vct{z}_\lambda=\lambda\vct{z}_1+\tau\sqrt{\lambda(1-\lambda)}\vct{\epsilon}$.
        \State Update $\theta=(\vct{m},\psi)$ using the averaged loss in Eq.~\eqref{eq:ram-loss}.
    \EndFor
\EndFor
\State Roll out a fresh endpoint $\vct{z}_1$ with the learned $\vct{b}_\theta$; return $\mathcal D(\vct{s}+\vct{z}_1)$.
\end{algorithmic}
\end{algorithm}

\subsection{Semantic Editing through a Model-Derived Reward}
\label{sec:method-edit}

An image--text score can increase without fulfilling a specific editing request. We therefore use the pretrained model's conditional information to define a semantic reward for the same endpoint solver.

Prior work observes that text conditioning shapes semantic content and spatial structure during early denoising~\citep{balaji2022ediffi,hertz2023p2p}. This motivates querying a higher-noise state $\vct{x}_t=(1-t)\vct{y}+t\vct{\epsilon}$ rather than only the clean candidate $\vct{y}$. We draw $t\sim\nu$ from a fixed high-noise interval inside $(0,1)$ and independent $\vct{\epsilon}\sim\Normal(0,\mat I)$. This direct construction avoids a pretrained sampling rollout.

Inspired by classifier-free guidance~\citep{ho2022cfg}, we compare target and source descriptions $c_{\mathrm{tgt}}$ and $c_{\mathrm{src}}$ at the same noisy candidate. Let $p_t(\cdot\mid c)$ be the noisy conditional density. Define
\begin{equation}
E_{\mathrm{edit}}(\vct{y})=\E_{t,\vct{\epsilon}}\!\left[\log\frac{p_t(\vct{x}_t\mid c_{\mathrm{tgt}})}{p_t(\vct{x}_t\mid c_{\mathrm{src}})}\right].
\label{eq:semantic-reward}
\end{equation}
Let $\vct{s}_t(\vct{x},c)=\nabla_{\vct{x}}\log p_t(\vct{x}\mid c)$ be the conditional score and $\vct{v}_c$ the conditional velocity. Under SD3's data-to-noise convention, their relation is
\begin{equation}
\label{eq:semantic-score-velocity}
\vct{s}_t(\vct{x},c)=-\frac{\vct{x}+(1-t)\vct{v}_c(\vct{x},t)}{t}.
\end{equation}
Taking the target-minus-source score difference cancels $\vct{x}_t$. The linear Jacobian $\partial\vct{x}_t/\partial\vct{y}=(1-t)\mat I$ then gives
\begin{equation}
\vct{g}_{\mathrm{edit}}(\vct{y}):=\nabla_{\vct{y}}E_{\mathrm{edit}}(\vct{y})
=\E_{t,\vct{\epsilon}}\!\left[\frac{(1-t)^2}{t}\bigl(\vct{v}_{c_{\mathrm{src}}}-\vct{v}_{c_{\mathrm{tgt}}}\bigr)(\vct{x}_t,t)\right].
\label{eq:semantic-gradient}
\end{equation}
Thus paired conditional forward predictions provide the gradient without backpropagation through the pretrained backbone; Appendix~\ref{app:edit-term} gives the derivation.

Adding the reward with weight $\eta\geq0$ subtracts $\eta E_{\mathrm{edit}}(\vct{s}+\vct{z})$ from the terminal cost, giving the practical label
\begin{equation}
\widehat{\vct{b}}_{\vct{s}}^{\mathrm{edit}}(\vct{z}_1)
=\widehat{\vct{b}}_{\vct{s}}(\vct{z}_1)
+\underbrace{\tau^2\eta}_{c_e}\widehat{\vct{g}}_{\mathrm{edit}}(\vct{s}+\vct{z}_1).
\label{eq:semantic-label}
\end{equation}
Here $\widehat{\vct{g}}_{\mathrm{edit}}$ estimates Eq.~\eqref{eq:semantic-gradient} using paired source/target model predictions. We cache and reuse the augmented label in the same bridge regression and refresh loop of Sec.~\ref{sec:method-solver}.

\section{Experiments}
\label{sec:experiments}

Following \citet{chang2026itoc}, we evaluate four reward-guided editing tasks with 300 source--condition pairs each. Human preference optimizes ImageReward~\citep{xu2023imagereward}. Style transfer minimizes the Gram-matrix difference between output and reference-image features from CLIP ViT-B/16~\citep{radford2021clip}. Counterfactual generation maximizes the target-class logit of a robust ResNet-50 classifier~\citep{engstrom2019robustness}, using a class label as the reward condition. Text-guided facial editing maximizes CLIP ViT-L/14 image--text similarity for the requested attribute. Each task also reports validation metrics not directly optimized: HPSv2~\citep{wu2023hpsv2}, CLIPScore, and Aesthetic for human preference; reference-image CLIP-I and DINO~\citep{caron2021dino} similarity for style; target-class text CLIPScore for counterfactual generation; and ImageReward and HPSv2 for facial editing. These metrics assess whether target improvements generalize beyond the optimized reward.

We use Stable Diffusion 3 Medium (SD3)~\citep{esser2024sd3} for the main quality comparison. The baselines are ITOC~\citep{chang2026itoc}, gradient ascent (GA), DPS~\citep{chung2023dps}, FreeDoM~\citep{yu2023freedom}, TFG~\citep{ye2024tfg}, and MPGD~\citep{he2024mpgd}; GA performs gradient ascent in latent space. Appendix~\ref{app:experiments} details the datasets, reward models, evaluation protocol, and baseline implementations, and provides the SD1.5 results.

\subsection{Main Results}
\label{sec:experiments-results}

FASTER achieves the best target metric in all four SD3 tasks and leads several validation metrics (Table~\ref{tab:main-sd3}). For human preference, the ImageReward improvement is accompanied by the highest HPSv2 and CLIPScore; for counterfactual generation, both the target-class logit and validation CLIPScore lead the comparison. Facial editing similarly combines the highest target CLIPScore with the best ImageReward. In style transfer, FASTER achieves the lowest GramDiff and highest reference CLIP-I, although its DINO similarity remains below the best result. These gains extend to multiple metrics that are not directly optimized, although FASTER does not lead every validation metric.


\begin{table}[!t]
\caption{Results on Stable Diffusion 3. Each task includes a target metric, which is directly optimized, and validation metrics used to assess potential reward hacking. Arrows indicate whether higher ($\uparrow$) or lower ($\downarrow$) values are better. Bold values indicate the best result for each metric.}
\label{tab:main-sd3}
\centering
\setlength{\tabcolsep}{3pt}
\resizebox{\linewidth}{!}{%
\begin{tabular}{lrrrrrrr}
\toprule
Metric & ITOC & GA & DPS & FreeDoM & TFG & MPGD & \textbf{FASTER} \\
\midrule
\rowcolor{taskband}
\multicolumn{8}{l}{\textbf{Human preference}} \\
{\scriptsize\color{brightorange}TARGET}\,\textbf{ImageReward}$\uparrow$ & 1.8326 & 1.8645 & 1.4687 & 1.6535 & 1.5128 & 0.9413 & \textbf{1.8958} \\
\hdashline[5pt/3pt]\noalign{\vskip 2pt}
{\scriptsize\color{softpurple}VALIDATION}\,HPSv2$\uparrow$ & 0.2619 & 0.2608 & 0.2579 & 0.2576 & 0.2580 & 0.2563 & \textbf{0.2624} \\
{\scriptsize\color{softpurple}VALIDATION}\,CLIPScore$\uparrow$ & 0.2734 & 0.2776 & 0.2566 & 0.2388 & 0.2539 & 0.2655 & \textbf{0.2783} \\
{\scriptsize\color{softpurple}VALIDATION}\,Aesthetic$\uparrow$ & \textbf{5.7933} & 5.6329 & 5.5978 & 5.6719 & 5.4951 & 5.4667 & 5.6991 \\
\midrule[.8pt]
\rowcolor{taskband}
\multicolumn{8}{l}{\textbf{Style transfer}} \\
{\scriptsize\color{brightorange}TARGET}\,\textbf{GramDiff}$\downarrow$ & 8.3341 & 9.6810 & 7.3644 & 6.9180 & 9.9923 & 10.2211 & \textbf{4.9746} \\
\hdashline[5pt/3pt]\noalign{\vskip 2pt}
{\scriptsize\color{softpurple}VALIDATION}\,CLIP-I$_{\mathrm{sty}}$$\uparrow$ & 0.5022 & 0.4951 & 0.5079 & 0.5068 & 0.5104 & 0.4945 & \textbf{0.5168} \\
{\scriptsize\color{softpurple}VALIDATION}\,DINO$_{\mathrm{sty}}$$\uparrow$ & 0.1727 & 0.1233 & 0.2361 & \textbf{0.2460} & 0.2013 & 0.1478 & 0.1442 \\
\midrule[.8pt]
\rowcolor{taskband}
\multicolumn{8}{l}{\textbf{Counterfactual generation}} \\
{\scriptsize\color{brightorange}TARGET}\,\textbf{Logit$_{\mathrm{tgt}}$}$\uparrow$ & 18.8789 & 11.4157 & 20.5937 & 21.6252 & 20.6566 & 16.0873 & \textbf{29.8948} \\
\hdashline[5pt/3pt]\noalign{\vskip 2pt}
{\scriptsize\color{softpurple}VALIDATION}\,CLIPScore$\uparrow$ & 0.1884 & 0.1620 & 0.1882 & 0.1950 & 0.1882 & 0.1823 & \textbf{0.2253} \\
\midrule[.8pt]
\rowcolor{taskband}
\multicolumn{8}{l}{\textbf{Text-guided facial editing}} \\
{\scriptsize\color{brightorange}TARGET}\,\textbf{CLIPScore}$\uparrow$ & 0.2654 & 0.2121 & 0.2544 & 0.2495 & 0.2509 & 0.2093 & \textbf{0.3538} \\
\hdashline[5pt/3pt]\noalign{\vskip 2pt}
{\scriptsize\color{softpurple}VALIDATION}\,ImageReward$\uparrow$ & -0.8159 & -0.8309 & -0.7610 & -0.7436 & -0.7539 & -0.9225 & \textbf{-0.6027} \\
{\scriptsize\color{softpurple}VALIDATION}\,HPSv2$\uparrow$ & \textbf{0.2573} & 0.2564 & 0.2531 & 0.2549 & 0.2535 & 0.2572 & 0.2533 \\
\bottomrule
\end{tabular}}
\end{table}

\begin{figure}[H]
\centering
\setlength{\abovecaptionskip}{0pt}
\includegraphics[width=\linewidth]{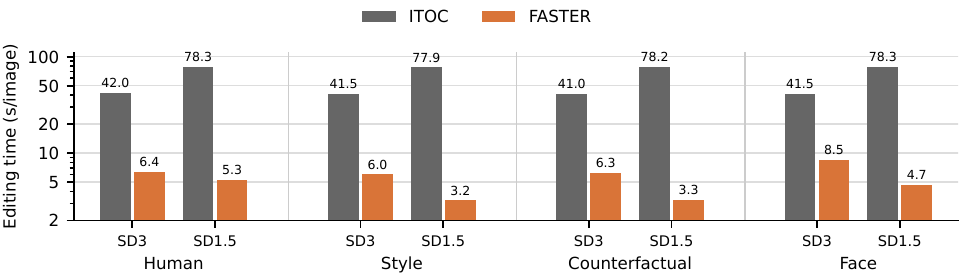}
\vspace{-1pt}
\caption{Editing time of ITOC and FASTER across four tasks on SD3 and SD1.5.}
\label{fig:itoc-faster-runtime}
\end{figure}

Alongside these quality results, FASTER offers a substantial editing-time advantage over the baseline that optimizes controls along pretrained generation trajectories (Fig.~\ref{fig:itoc-faster-runtime}). FASTER achieves speedups of up to $6.91\times$ on SD3 and $24.14\times$ on SD1.5. This advantage is consistent with assigning repeated updates to a small network and reusing costly endpoint feedback. A $768\times768$ SD3 ImageReward edit takes approximately 7.7\,s of editing time on a single consumer GPU. Appendix~\ref{app:experiments} gives the timing protocol, memory measurements, and comparisons with all baselines.

\vspace{-12pt}

\begin{figure}[H]
\centering
\makebox[\linewidth][c]{\includegraphics[width=1.0\linewidth]{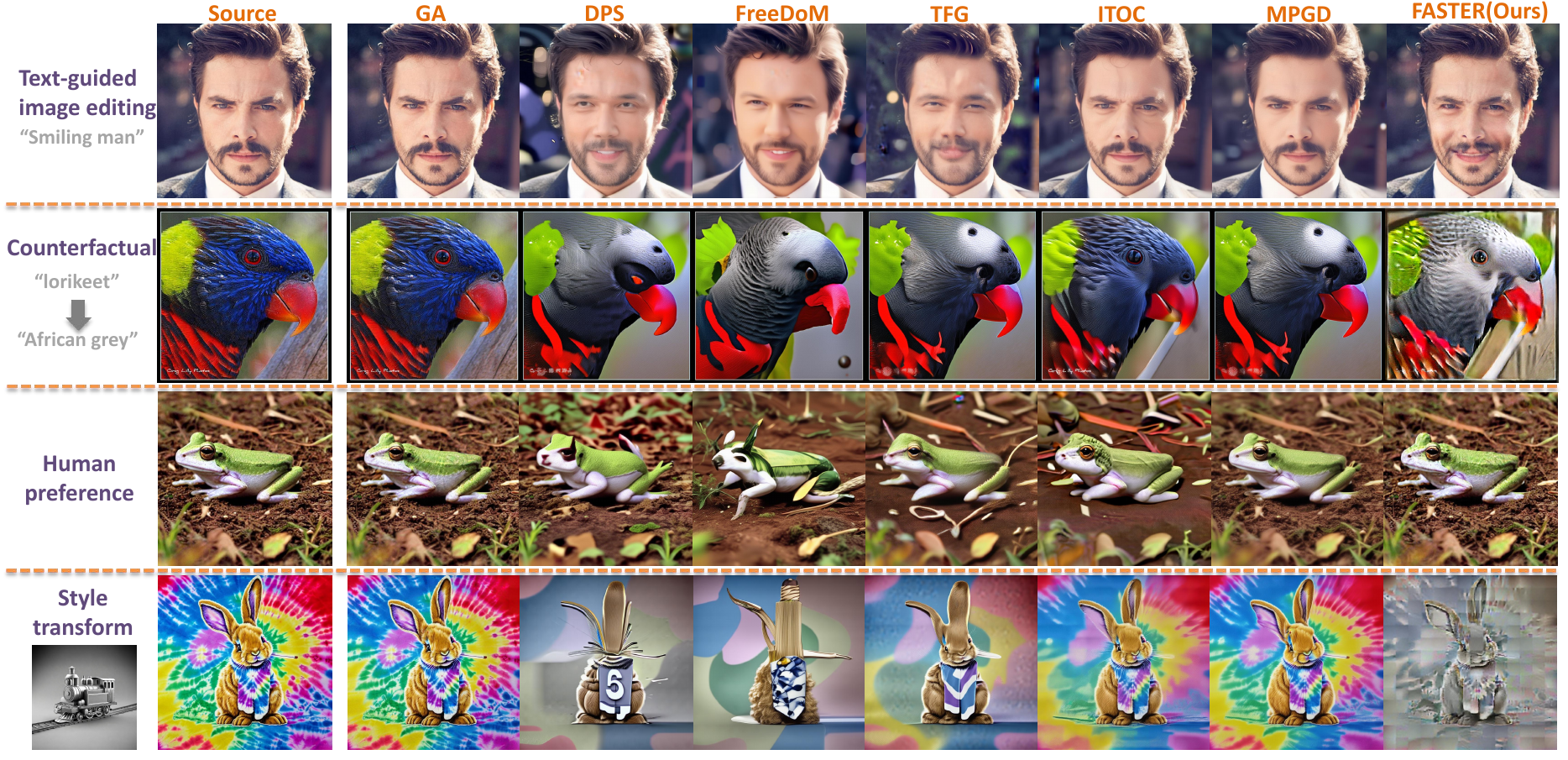}}
\vspace{-1pt}
\caption{Qualitative comparisons on four SD3 editing tasks. Rows show text-guided image editing, counterfactual generation, human preference enhancement, and style transfer. For text-guided image editing, the target prompt is supplied to the pretrained generative model of each applicable method. The style reference is shown on the left.}
\label{fig:sd3-qualitative}
\end{figure}

The qualitative examples in Fig.~\ref{fig:sd3-qualitative} show how these edits retain source content while changing the requested attributes. In the text-guided example, it produces a smile while largely preserving the man's facial appearance, hairstyle, and composition. The counterfactual example changes the lorikeet toward an African grey appearance while retaining the source bird's head pose and outline. In the human-preference example, the frog and surrounding scene remain recognizable. For style transfer, the rabbit retains its overall silhouette and pose while acquiring the silver-grey palette and metallic appearance of the reference train. Together, these examples illustrate targeted changes without wholesale replacement of the source composition.

\vspace{-10pt}
\FloatBarrier

\subsection{Editing beyond CLIPScore}
\label{sec:experiments-edit-term}

CLIPScore can assess image--text alignment, but maximizing it does not guarantee the requested semantic edit.

In Fig.~\ref{fig:edit-term}, optimization raises CLIPScore from $0.223$ to $0.356$ without adding sunglasses. The Editing Term (Sec.~\ref{sec:method-edit}) produces visible sunglasses despite a lower score of $0.271$, supplying semantic guidance beyond score maximization. Further Editing Term examples and parameter sweeps appear in Appendix~\ref{app:edit-term}.

\par\smallskip
\noindent\begin{minipage}{\linewidth}
\centering
\includegraphics[width=0.70\linewidth]{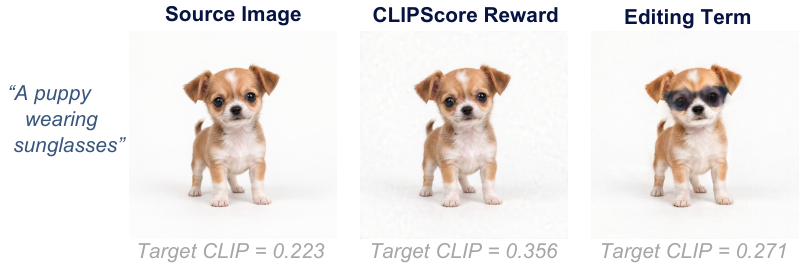}
\setlength{\abovecaptionskip}{3pt}
\captionof{figure}{FASTER with CLIPScore optimization or the Editing Term for the same sunglasses request. Values are CLIPScore.}
\label{fig:edit-term}
\end{minipage}
\par\medskip

\subsection{Reward strength and outer-loop budget}
\label{sec:experiments-k}
\begin{figure}[H]
\centering
\includegraphics[width=\linewidth]{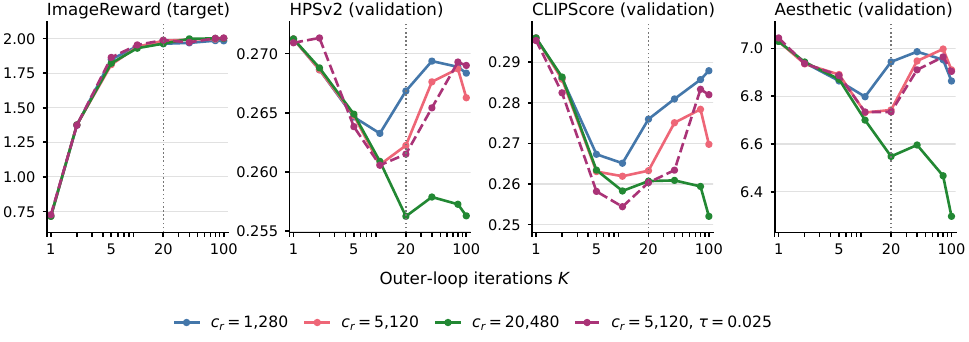}
\vspace{-10pt}
\caption{Effect of $K$ and four parameter settings on the SD3 ImageReward task. Each curve averages the same three source--prompt pairs at saved checkpoints from one run per setting. All settings use $\alpha\tau^2=1$ and $c_p=0$. Solid curves use $\tau=0.0125$; the dashed curve uses $\tau=0.025$. The dotted vertical line marks $K=20$.}
\label{fig:human-k-ablation}
\end{figure}

Figure~\ref{fig:human-k-ablation} varies the outer-loop budget $K$ and reward coefficient $c_r$ on the ImageReward task. For the tested source--prompt pairs, a suitable $c_r$ allows more outer-loop steps to improve ImageReward while validation metrics recover or remain strong at later checkpoints. In contrast, the largest $c_r$ brings little additional ImageReward over the moderate settings, but HPSv2, CLIPScore, and Aesthetic all deteriorate as $K$ grows. The gap between a saturated target reward and worsening validation metrics indicates reward hacking in these settings. These observations show that the benefit of increasing $K$ depends on the reward strength. Appendix~\ref{app:edit-term} also shows the visual progression with $K$ when the Editing Term drives a larger semantic change.

\vspace{-8pt}

\FloatBarrier

\section{Conclusion}
\label{sec:conclusion}

We introduced a decoupled view of reward-guided image editing at test time, in which pretrained models provide visual knowledge while a separate lightweight process executes the edit. Our theoretical framework jointly accounts for reward improvement, source preservation, and prior preferences, establishing a basis for specifying desirable outputs independently of pretrained generation trajectories. Building on this framework, FASTER trains a small per-image network and reuses candidate outputs and their costly feedback across multiple inexpensive updates. A complementary editing term obtains semantic directions from conditional model predictions without backpropagating through the pretrained backbone, supporting changes that image--text score maximization alone may fail to produce. Across four tasks, FASTER achieves competitive target and independent validation metrics, with editing-time speedups of up to $6.91\times$ on SD3 and $24.14\times$ on SD1.5 relative to the evaluated baseline that optimizes controls along pretrained generation trajectories.



\section*{Acknowledgments}
This work was supported by the National Natural Science Foundation of China under Grant 62406195, together with support from the HPC Platform of ShanghaiTech University, the MoE Key Laboratory of Intelligent Perception and Human-Machine Collaboration (ShanghaiTech University), and the Shanghai Engineering Research Center of Intelligent Vision and Imaging. Computational resources for this work were also provided in part by Fcloud CO., LTD.
\bibliographystyle{iclr2027_conference}
\bibliography{references}

\begin{thebibliography}{53}
\providecommand{\natexlab}[1]{#1}
\providecommand{\url}[1]{\texttt{#1}}
\expandafter\ifx\csname urlstyle\endcsname\relax
  \providecommand{\doi}[1]{doi: #1}\else
  \providecommand{\doi}{doi: \begingroup \urlstyle{rm}\Url}\fi

\bibitem[Balaji et~al.(2022)Balaji, Nah, Huang, Vahdat, Song, Zhang, Kreis, Aittala, Aila, Laine, Catanzaro, Karras, and Liu]{balaji2022ediffi}
Yogesh Balaji, Seungjun Nah, Xun Huang, Arash Vahdat, Jiaming Song, Qinsheng Zhang, Karsten Kreis, Miika Aittala, Timo Aila, Samuli Laine, Bryan Catanzaro, Tero Karras, and Ming-Yu Liu.
\newblock {eDiff-I}: Text-to-image diffusion models with an ensemble of expert denoisers.
\newblock \emph{arXiv preprint arXiv:2211.01324}, 2022.
\newblock URL \url{https://arxiv.org/abs/2211.01324}.

\bibitem[Bansal et~al.(2023)Bansal, Chu, Schwarzschild, Sengupta, Goldblum, Geiping, and Goldstein]{bansal2023universal}
Arpit Bansal, Hong-Min Chu, Avi Schwarzschild, Soumyadip Sengupta, Micah Goldblum, Jonas Geiping, and Tom Goldstein.
\newblock Universal guidance for diffusion models.
\newblock In \emph{Proceedings of the IEEE/CVF Conference on Computer Vision and Pattern Recognition (CVPR) Workshops}, pp.\  843--852, 2023.
\newblock URL \url{https://openaccess.thecvf.com/content/CVPR2023W/GCV/html/Bansal_Universal_Guidance_for_Diffusion_Models_CVPRW_2023_paper.html}.

\bibitem[Ben-Hamu et~al.(2024)Ben-Hamu, Puny, Gat, Karrer, Singer, and Lipman]{benhamu2024dflow}
Heli Ben-Hamu, Omri Puny, Itai Gat, Brian Karrer, Uriel Singer, and Yaron Lipman.
\newblock {D-Flow}: Differentiating through flows for controlled generation.
\newblock In \emph{Forty-first International Conference on Machine Learning}, volume 235, pp.\  3462--3483. PMLR, 2024.
\newblock URL \url{https://proceedings.mlr.press/v235/ben-hamu24a.html}.

\bibitem[Cao et~al.(2023)Cao, Wang, Qi, Shan, Qie, and Zheng]{cao2023masactrl}
Mingdeng Cao, Xintao Wang, Zhongang Qi, Ying Shan, Xiaohu Qie, and Yinqiang Zheng.
\newblock {MasaCtrl}: Tuning-free mutual self-attention control for consistent image synthesis and editing.
\newblock In \emph{2023 IEEE/CVF International Conference on Computer Vision (ICCV)}, pp.\  22503--22513. IEEE, 2023.

\bibitem[Caron et~al.(2021)Caron, Touvron, Misra, J{\'e}gou, Mairal, Bojanowski, and Joulin]{caron2021dino}
Mathilde Caron, Hugo Touvron, Ishan Misra, Herv{\'e} J{\'e}gou, Julien Mairal, Piotr Bojanowski, and Armand Joulin.
\newblock Emerging properties in self-supervised vision transformers.
\newblock In \emph{Proceedings of the IEEE/CVF International Conference on Computer Vision}, 2021.
\newblock URL \url{https://arxiv.org/abs/2104.14294}.

\bibitem[Chang et~al.(2026)Chang, Kim, and Ye]{chang2026itoc}
Jinho Chang, Jaemin Kim, and Jong~Chul Ye.
\newblock Training-free reward-guided image editing via trajectory optimal control.
\newblock In \emph{The Fourteenth International Conference on Learning Representations}, 2026.
\newblock URL \url{https://openreview.net/forum?id=YL2NgqN3Vh}.

\bibitem[Chung et~al.(2023)Chung, Kim, Mccann, Klasky, and Ye]{chung2023dps}
Hyungjin Chung, Jeongsol Kim, Michael~Thompson Mccann, Marc~Louis Klasky, and Jong~Chul Ye.
\newblock Diffusion posterior sampling for general noisy inverse problems.
\newblock In \emph{The Eleventh International Conference on Learning Representations}, 2023.
\newblock URL \url{https://openreview.net/forum?id=OnD9zGAGT0k}.

\bibitem[Deng et~al.(2009)Deng, Dong, Socher, Li, Li, and Fei-Fei]{deng2009imagenet}
Jia Deng, Wei Dong, Richard Socher, Li-Jia Li, Kai Li, and Li~Fei-Fei.
\newblock {ImageNet}: A large-scale hierarchical image database.
\newblock In \emph{Proceedings of the IEEE Conference on Computer Vision and Pattern Recognition}, pp.\  248--255, 2009.
\newblock \doi{10.1109/CVPR.2009.5206848}.

\bibitem[Domingo-Enrich et~al.(2025)Domingo-Enrich, Drozdzal, Karrer, and Chen]{domingoenrich2025adjoint}
Carles Domingo-Enrich, Michal Drozdzal, Brian Karrer, and Ricky T.~Q. Chen.
\newblock Adjoint matching: Fine-tuning flow and diffusion generative models with memoryless stochastic optimal control.
\newblock In \emph{International Conference on Learning Representations}, 2025.
\newblock URL \url{https://openreview.net/forum?id=xQBRrtQM8u}.

\bibitem[Engstrom et~al.(2019)Engstrom, Ilyas, Salman, Santurkar, and Tsipras]{engstrom2019robustness}
Logan Engstrom, Andrew Ilyas, Hadi Salman, Shibani Santurkar, and Dimitris Tsipras.
\newblock Robustness ({Python} library), 2019.
\newblock URL \url{https://github.com/MadryLab/robustness}.

\bibitem[Esser et~al.(2024)Esser, Kulal, Blattmann, Entezari, M{\"u}ller, Saini, Levi, Lorenz, Sauer, Boesel, Podell, Dockhorn, English, and Rombach]{esser2024sd3}
Patrick Esser, Sumith Kulal, Andreas Blattmann, Rahim Entezari, Jonas M{\"u}ller, Harry Saini, Yam Levi, Dominik Lorenz, Axel Sauer, Frederic Boesel, Dustin Podell, Tim Dockhorn, Zion English, and Robin Rombach.
\newblock Scaling rectified flow transformers for high-resolution image synthesis.
\newblock In \emph{Proceedings of the 41st International Conference on Machine Learning}, volume 235, pp.\  12606--12633. PMLR, 2024.
\newblock URL \url{https://proceedings.mlr.press/v235/esser24a.html}.

\bibitem[Eyring et~al.(2024)Eyring, Karthik, Roth, Dosovitskiy, and Akata]{eyring2024reno}
Luca Eyring, Shyamgopal Karthik, Karsten Roth, Alexey Dosovitskiy, and Zeynep Akata.
\newblock {ReNO}: Enhancing one-step text-to-image models through reward-based noise optimization.
\newblock In \emph{Advances in Neural Information Processing Systems}, volume~37, pp.\  125487--125519. Curran Associates, Inc., 2024.

\bibitem[Feng \& Bouman(2024)Feng and Bouman]{feng2024surrogate}
Berthy Feng and Katherine Bouman.
\newblock Variational bayesian imaging with an efficient surrogate score-based prior.
\newblock \emph{Transactions on Machine Learning Research}, 2024.
\newblock ISSN 2835-8856.
\newblock URL \url{https://openreview.net/forum?id=db2pFKVcm1}.

\bibitem[Feng et~al.(2023)Feng, Smith, Rubinstein, Chang, Bouman, and Freeman]{feng2023priors}
Berthy~T Feng, Jamie Smith, Michael Rubinstein, Huiwen Chang, Katherine~L Bouman, and William~T Freeman.
\newblock Score-based diffusion models as principled priors for inverse imaging.
\newblock In \emph{2023 IEEE/CVF International Conference on Computer Vision (ICCV)}, pp.\  10486--10497. IEEE, 2023.

\bibitem[Havens et~al.(2025)Havens, Miller, Yan, Domingo-Enrich, Sriram, Levine, Wood, Hu, Amos, Karrer, Fu, Liu, and Chen]{havens2025adjointsampling}
Aaron~J Havens, Benjamin~Kurt Miller, Bing Yan, Carles Domingo-Enrich, Anuroop Sriram, Daniel~S. Levine, Brandon~M Wood, Bin Hu, Brandon Amos, Brian Karrer, Xiang Fu, Guan-Horng Liu, and Ricky T.~Q. Chen.
\newblock Adjoint sampling: Highly scalable diffusion samplers via adjoint matching.
\newblock In \emph{Forty-second International Conference on Machine Learning}, 2025.
\newblock URL \url{https://openreview.net/forum?id=6Eg1OrHmg2}.

\bibitem[He et~al.(2024)He, Murata, Lai, Takida, Uesaka, Kim, Liao, Mitsufuji, Kolter, Salakhutdinov, and Ermon]{he2024mpgd}
Yutong He, Naoki Murata, Chieh-Hsin Lai, Yuhta Takida, Toshimitsu Uesaka, Dongjun Kim, Wei-Hsiang Liao, Yuki Mitsufuji, J.~Zico Kolter, Ruslan Salakhutdinov, and Stefano Ermon.
\newblock Manifold preserving guided diffusion.
\newblock In \emph{International Conference on Learning Representations}, 2024.

\bibitem[Hertz et~al.(2023{\natexlab{a}})Hertz, Aberman, and Cohen-Or]{hertz2023dds}
Amir Hertz, Kfir Aberman, and Daniel Cohen-Or.
\newblock Delta denoising score.
\newblock In \emph{2023 IEEE/CVF International Conference on Computer Vision (ICCV)}, pp.\  2328--2337. IEEE, 2023{\natexlab{a}}.

\bibitem[Hertz et~al.(2023{\natexlab{b}})Hertz, Mokady, Tenenbaum, Aberman, Pritch, and Cohen-Or]{hertz2023p2p}
Amir Hertz, Ron Mokady, Jay Tenenbaum, Kfir Aberman, Yael Pritch, and Daniel Cohen-Or.
\newblock Prompt-to-prompt image editing with cross-attention control.
\newblock In \emph{The Eleventh International Conference on Learning Representations}, 2023{\natexlab{b}}.
\newblock URL \url{https://openreview.net/forum?id=_CDixzkzeyb}.

\bibitem[Hessel et~al.(2021)Hessel, Holtzman, Forbes, Le~Bras, and Choi]{hessel2021clipscore}
Jack Hessel, Ari Holtzman, Maxwell Forbes, Ronan Le~Bras, and Yejin Choi.
\newblock {CLIPScore}: A reference-free evaluation metric for image captioning.
\newblock In \emph{Proceedings of the 2021 Conference on Empirical Methods in Natural Language Processing}, pp.\  7514--7528, 2021.
\newblock URL \url{https://aclanthology.org/2021.emnlp-main.595/}.

\bibitem[Ho \& Salimans(2022)Ho and Salimans]{ho2022cfg}
Jonathan Ho and Tim Salimans.
\newblock Classifier-free diffusion guidance.
\newblock \emph{arXiv preprint arXiv:2207.12598}, 2022.
\newblock URL \url{https://arxiv.org/abs/2207.12598}.

\bibitem[Ho et~al.(2020)Ho, Jain, and Abbeel]{ho2020ddpm}
Jonathan Ho, Ajay Jain, and Pieter Abbeel.
\newblock Denoising diffusion probabilistic models.
\newblock In \emph{Advances in Neural Information Processing Systems}, volume~33, pp.\  6840--6851. Curran Associates, Inc., 2020.

\bibitem[Huang et~al.(2026)Huang, Lin, Shah, Nair, and Boffi]{huang2026fmrg}
Jerry~Y Huang, Justin Lin, Sheel Shah, Kartik Nair, and Nicholas~M Boffi.
\newblock How to guide your flow: Few-step alignment via flow map reward guidance.
\newblock In \emph{Proceedings of the 43rd International Conference on Machine Learning}, 2026.
\newblock URL \url{https://github.com/jrrhuang/fmrg}.

\bibitem[Karras et~al.(2018)Karras, Aila, Laine, and Lehtinen]{karras2018progressive}
Tero Karras, Timo Aila, Samuli Laine, and Jaakko Lehtinen.
\newblock Progressive growing of {GANs} for improved quality, stability, and variation.
\newblock In \emph{International Conference on Learning Representations}, 2018.
\newblock URL \url{https://arxiv.org/abs/1710.10196}.

\bibitem[Kim et~al.(2025)Kim, Kim, and Park]{kim2025das}
Sunwoo Kim, Minkyu Kim, and Dongmin Park.
\newblock Test-time alignment of diffusion models without reward over-optimization.
\newblock In \emph{The Thirteenth International Conference on Learning Representations}, 2025.
\newblock URL \url{https://openreview.net/forum?id=vi3DjUhFVm}.

\bibitem[Kirstain et~al.(2023)Kirstain, Polyak, Singer, Matiana, Penna, and Levy]{kirstain2023pickapic}
Yuval Kirstain, Adam Polyak, Uriel Singer, Shahbuland Matiana, Joe Penna, and Omer Levy.
\newblock {Pick-a-Pic}: An open dataset of user preferences for text-to-image generation.
\newblock In \emph{Advances in Neural Information Processing Systems}, volume~36, pp.\  36652--36663. Curran Associates, Inc., 2023.
\newblock URL \url{https://proceedings.neurips.cc/paper_files/paper/2023/hash/73aacd8b3b05b4b503d58310b523553c-Abstract-Conference.html}.

\bibitem[Kulikov et~al.(2025)Kulikov, Kleiner, Huberman-Spiegelglas, and Michaeli]{kulikov2025flowedit}
Vladimir Kulikov, Matan Kleiner, Inbar Huberman-Spiegelglas, and Tomer Michaeli.
\newblock {FlowEdit}: Inversion-free text-based editing using pre-trained flow models.
\newblock In \emph{2025 IEEE/CVF International Conference on Computer Vision (ICCV)}, pp.\  19721--19730. IEEE, 2025.

\bibitem[Lipman et~al.(2023)Lipman, Chen, Ben-Hamu, Nickel, and Le]{lipman2023flow}
Yaron Lipman, Ricky T.~Q. Chen, Heli Ben-Hamu, Maximilian Nickel, and Matthew Le.
\newblock Flow matching for generative modeling.
\newblock In \emph{The Eleventh International Conference on Learning Representations}, 2023.
\newblock URL \url{https://openreview.net/forum?id=PqvMRDCJT9t}.

\bibitem[Liu et~al.(2023)Liu, Gong, and Liu]{liu2023rectifiedflow}
Xingchao Liu, Chengyue Gong, and Qiang Liu.
\newblock Flow straight and fast: Learning to generate and transfer data with rectified flow.
\newblock In \emph{International Conference on Learning Representations}, 2023.

\bibitem[Mardani et~al.(2024)Mardani, Song, Kautz, and Vahdat]{mardani2024reddiff}
Morteza Mardani, Jiaming Song, Jan Kautz, and Arash Vahdat.
\newblock A variational perspective on solving inverse problems with diffusion models.
\newblock In \emph{International Conference on Learning Representations}, 2024.

\bibitem[Meng et~al.(2022)Meng, He, Song, Song, Wu, Zhu, and Ermon]{meng2022sdedit}
Chenlin Meng, Yutong He, Yang Song, Jiaming Song, Jiajun Wu, Jun-Yan Zhu, and Stefano Ermon.
\newblock {SDE}dit: Guided image synthesis and editing with stochastic differential equations.
\newblock In \emph{International Conference on Learning Representations}, 2022.

\bibitem[Miller(1995)]{miller1995wordnet}
George~A. Miller.
\newblock {WordNet}: A lexical database for {English}.
\newblock \emph{Communications of the ACM}, 38\penalty0 (11):\penalty0 39--41, 1995.
\newblock \doi{10.1145/219717.219748}.

\bibitem[Mokady et~al.(2023)Mokady, Hertz, Aberman, Pritch, and Cohen-Or]{mokady2023nti}
Ron Mokady, Amir Hertz, Kfir Aberman, Yael Pritch, and Daniel Cohen-Or.
\newblock Null-text inversion for editing real images using guided diffusion models.
\newblock In \emph{2023 IEEE/CVF Conference on Computer Vision and Pattern Recognition (CVPR)}, pp.\  6038--6047. IEEE, 2023.

\bibitem[Pan et~al.(2026)Pan, Zhu, Ma, Fu, Yu, Wang, and Shi]{pan2025unidbplusplus}
Mokai Pan, Kaizhen Zhu, Yuexin Ma, Yanwei Fu, Jingyi Yu, Jingya Wang, and Ye~Shi.
\newblock A unified and fast-sampling diffusion bridge framework via stochastic optimal control.
\newblock \emph{IEEE Transactions on Pattern Analysis and Machine Intelligence}, 2026.
\newblock \doi{10.1109/TPAMI.2026.3710696}.
\newblock URL \url{https://doi.org/10.1109/TPAMI.2026.3710696}.

\bibitem[Patel et~al.(2025)Patel, Wen, Metaxas, and Yang]{patel2025flowchef}
Maitreya Patel, Song Wen, Dimitris~N Metaxas, and Yezhou Yang.
\newblock {FlowChef}: Steering of rectified flow models for controlled generations.
\newblock In \emph{2025 IEEE/CVF International Conference on Computer Vision (ICCV)}, pp.\  15308--15318. IEEE, 2025.

\bibitem[Poole et~al.(2023)Poole, Jain, Barron, and Mildenhall]{poole2023dreamfusion}
Ben Poole, Ajay Jain, Jonathan~T. Barron, and Ben Mildenhall.
\newblock {DreamFusion}: Text-to-{3D} using {2D} diffusion.
\newblock In \emph{The Eleventh International Conference on Learning Representations}, 2023.
\newblock URL \url{https://openreview.net/forum?id=FjNys5c7VyY}.

\bibitem[Radford et~al.(2021)Radford, Kim, Hallacy, Ramesh, Goh, Agarwal, Sastry, Askell, Mishkin, Clark, Krueger, and Sutskever]{radford2021clip}
Alec Radford, Jong~Wook Kim, Chris Hallacy, Aditya Ramesh, Gabriel Goh, Sandhini Agarwal, Girish Sastry, Amanda Askell, Pamela Mishkin, Jack Clark, Gretchen Krueger, and Ilya Sutskever.
\newblock Learning transferable visual models from natural language supervision.
\newblock In \emph{International Conference on Machine Learning}, volume 139, pp.\  8748--8763. PMLR, 2021.
\newblock URL \url{https://proceedings.mlr.press/v139/radford21a.html}.

\bibitem[Rombach et~al.(2022)Rombach, Blattmann, Lorenz, Esser, and Ommer]{rombach2022ldm}
Robin Rombach, Andreas Blattmann, Dominik Lorenz, Patrick Esser, and Bj{\"o}rn Ommer.
\newblock High-resolution image synthesis with latent diffusion models.
\newblock In \emph{2022 IEEE/CVF Conference on Computer Vision and Pattern Recognition (CVPR)}, pp.\  10674--10685. IEEE, 2022.

\bibitem[Shin et~al.(2026)Shin, Shin, Zhu, Guo, Chen, Lee, Choi, and Choi]{shin2026eam}
Jeongwoo Shin, Dongsoo Shin, Yuchen Zhu, Wei Guo, Yongxin Chen, Joonseok Lee, Jaewoong Choi, and Jaemoo Choi.
\newblock Efficient adjoint matching for fine-tuning diffusion models.
\newblock \emph{arXiv preprint arXiv:2605.11480}, 2026.

\bibitem[Singhal et~al.(2025)Singhal, Horvitz, Teehan, Ren, Yu, McKeown, and Ranganath]{singhal2025fk}
Raghav Singhal, Zachary Horvitz, Ryan Teehan, Mengye Ren, Zhou Yu, Kathleen McKeown, and Rajesh Ranganath.
\newblock A general framework for inference-time scaling and steering of diffusion models.
\newblock In \emph{Forty-second International Conference on Machine Learning}, 2025.
\newblock URL \url{https://openreview.net/forum?id=Jp988ELppQ}.

\bibitem[Song et~al.(2021)Song, Sohl-Dickstein, Kingma, Kumar, Ermon, and Poole]{song2020score}
Yang Song, Jascha Sohl-Dickstein, Diederik~P. Kingma, Abhishek Kumar, Stefano Ermon, and Ben Poole.
\newblock Score-based generative modeling through stochastic differential equations.
\newblock In \emph{International Conference on Learning Representations}, 2021.

\bibitem[Susladkar et~al.(2026)Susladkar, Jang, Prakash, Juvekar, Shah, Barik, Bashir, Wahed, Shrirao, and Lourentzou]{susladkar2026rewardflow}
Onkar Susladkar, Dong-Hwan Jang, Tushar Prakash, Adheesh Juvekar, Vedant Shah, Ayush Barik, Nabeel Bashir, Muntasir Wahed, Ritish Shrirao, and Ismini Lourentzou.
\newblock {RewardFlow}: Generate images by optimizing what you reward.
\newblock In \emph{Proceedings of the IEEE/CVF Conference on Computer Vision and Pattern Recognition (CVPR)}, pp.\  20066--20076, June 2026.

\bibitem[Tang et~al.(2025)Tang, Peng, Tang, Hong, Wang, and Chang]{tang2025dno}
Zhiwei Tang, Jiangweizhi Peng, Jiasheng Tang, Mingyi Hong, Fan Wang, and Tsung-Hui Chang.
\newblock Inference-time alignment of diffusion models with direct noise optimization.
\newblock In \emph{Forty-second International Conference on Machine Learning}, 2025.
\newblock URL \url{https://openreview.net/forum?id=JpbqiD7n9r}.

\bibitem[Tumanyan et~al.(2023)Tumanyan, Geyer, Bagon, and Dekel]{tumanyan2023pnp}
Narek Tumanyan, Michal Geyer, Shai Bagon, and Tali Dekel.
\newblock Plug-and-play diffusion features for text-driven image-to-image translation.
\newblock In \emph{2023 IEEE/CVF Conference on Computer Vision and Pattern Recognition (CVPR)}, pp.\  1921--1930. IEEE, 2023.

\bibitem[Wallace et~al.(2023)Wallace, Gokul, Ermon, and Naik]{wallace2023doodl}
Bram Wallace, Akash Gokul, Stefano Ermon, and Nikhil Naik.
\newblock End-to-end diffusion latent optimization improves classifier guidance.
\newblock In \emph{2023 IEEE/CVF International Conference on Computer Vision (ICCV)}, pp.\  7246--7256. IEEE, 2023.

\bibitem[Wang et~al.(2023)Wang, Lu, Wang, Bao, Li, Su, and Zhu]{wang2023prolificdreamer}
Zhengyi Wang, Cheng Lu, Yikai Wang, Fan Bao, Chongxuan Li, Hang Su, and Jun Zhu.
\newblock {ProlificDreamer}: High-fidelity and diverse text-to-{3D} generation with variational score distillation.
\newblock In \emph{Advances in Neural Information Processing Systems}, volume~36, pp.\  8406--8441. Curran Associates, Inc., 2023.

\bibitem[Wu et~al.(2023)Wu, Hao, Sun, Chen, Zhu, Zhao, and Li]{wu2023hpsv2}
Xiaoshi Wu, Yiming Hao, Keqiang Sun, Yixiong Chen, Feng Zhu, Rui Zhao, and Hongsheng Li.
\newblock Human preference score v2: A solid benchmark for evaluating human preferences of text-to-image synthesis.
\newblock \emph{arXiv preprint arXiv:2306.09341}, 2023.
\newblock URL \url{https://arxiv.org/abs/2306.09341}.

\bibitem[Xu et~al.(2023)Xu, Liu, Wu, Tong, Li, Ding, Tang, and Dong]{xu2023imagereward}
Jiazheng Xu, Xiao Liu, Yuchen Wu, Yuxuan Tong, Qinkai Li, Ming Ding, Jie Tang, and Yuxiao Dong.
\newblock {ImageReward}: Learning and evaluating human preferences for text-to-image generation.
\newblock In \emph{Advances in Neural Information Processing Systems}, volume~36, pp.\  15903--15935. Curran Associates, Inc., 2023.
\newblock URL \url{https://proceedings.neurips.cc/paper_files/paper/2023/hash/33646ef0ed554145eab65f6250fab0c9-Abstract-Conference.html}.

\bibitem[Xu et~al.(2024)Xu, Huang, Pan, Ma, and Chai]{xu2024infedit}
Sihan Xu, Yidong Huang, Jiayi Pan, Ziqiao Ma, and Joyce Chai.
\newblock Inversion-free image editing with language-guided diffusion models.
\newblock In \emph{Proceedings of the IEEE/CVF Conference on Computer Vision and Pattern Recognition (CVPR)}, pp.\  9452--9461, June 2024.

\bibitem[Yang et~al.(2024)Yang, Ding, Cai, Yu, Wang, and Shi]{yang2024dsg}
Lingxiao Yang, Shutong Ding, Yifan Cai, Jingyi Yu, Jingya Wang, and Ye~Shi.
\newblock Guidance with spherical gaussian constraint for conditional diffusion.
\newblock In \emph{Forty-first International Conference on Machine Learning}, volume 235, pp.\  56071--56095. PMLR, 2024.
\newblock URL \url{https://proceedings.mlr.press/v235/yang24h.html}.

\bibitem[Ye et~al.(2024)Ye, Lin, Han, Xu, Liu, Liang, Ma, Zou, and Ermon]{ye2024tfg}
Haotian Ye, Haowei Lin, Jiaqi Han, Minkai Xu, Sheng Liu, Yitao Liang, Jianzhu Ma, James Zou, and Stefano Ermon.
\newblock {TFG}: Unified training-free guidance for diffusion models.
\newblock In \emph{Advances in Neural Information Processing Systems}, volume~37, pp.\  22370--22417. Curran Associates, Inc., 2024.

\bibitem[Yu et~al.(2023)Yu, Wang, Zhao, Ghanem, and Zhang]{yu2023freedom}
Jiwen Yu, Yinhuai Wang, Chen Zhao, Bernard Ghanem, and Jian Zhang.
\newblock {FreeDoM}: Training-free energy-guided conditional diffusion model.
\newblock In \emph{2023 IEEE/CVF International Conference on Computer Vision (ICCV)}, pp.\  23117--23127. IEEE, 2023.

\bibitem[Zhang et~al.(2018)Zhang, Isola, Efros, Shechtman, and Wang]{zhang2018lpips}
Richard Zhang, Phillip Isola, Alexei~A. Efros, Eli Shechtman, and Oliver Wang.
\newblock The unreasonable effectiveness of deep features as a perceptual metric.
\newblock In \emph{Proceedings of the IEEE Conference on Computer Vision and Pattern Recognition}, pp.\  586--595, 2018.
\newblock URL \url{https://arxiv.org/abs/1801.03924}.

\bibitem[Zhu et~al.(2025)Zhu, Pan, Ma, Fu, Yu, Wang, and Shi]{zhu2025unidb}
Kaizhen Zhu, Mokai Pan, Yuexin Ma, Yanwei Fu, Jingyi Yu, Jingya Wang, and Ye~Shi.
\newblock {UniDB}: A unified diffusion bridge framework via stochastic optimal control.
\newblock In \emph{Proceedings of the 42nd International Conference on Machine Learning}, volume 267, pp.\  79985--80012. PMLR, 2025.
\newblock URL \url{https://proceedings.mlr.press/v267/zhu25o.html}.

\end{thebibliography}
\clearpage
\appendix
\section{Source-Conditioned Target: Derivation and Scope}
\label{app:conditional}
\subsection{Per-image variational objective}
For a fixed source $\vct{s}$, consider $F_{\vct{s}}$ in Eq.~\eqref{eq:editing-objective}. Assume $p_{\mathrm{pre}}$ is a density and the following normalizer is finite and positive:
\begin{subequations}
\begin{align}
Z_{\vct{s}}^{(1)}&=\int p_{\mathrm{pre}}(\vct{y})\exp[-\alpha D(\vct{s},\vct{y})+\beta r(\vct{y})]d\vct{y},\\
\pi_{\vct{s}}^{(1)}(\vct{y})&=(Z_{\vct{s}}^{(1)})^{-1}p_{\mathrm{pre}}(\vct{y})\exp[-\alpha D(\vct{s},\vct{y})+\beta r(\vct{y})].
\end{align}
\end{subequations}
For densities with finite objective, substitution gives
\begin{subequations}
\begin{align}
\KL(\rho\Vert\pi_{\vct{s}}^{(1)})
&=\KL(\rho\Vert p_{\mathrm{pre}})+\alpha\E_\rho D(\vct{s},Y)-\beta\E_\rho r(Y)+\log Z_{\vct{s}}^{(1)},\\
F_{\vct{s}}(\rho)&=\KL(\rho\Vert\pi_{\vct{s}}^{(1)})-\log Z_{\vct{s}}^{(1)}.
\end{align}
\end{subequations}
Nonnegativity of KL proves the unique minimizer, up to equality almost everywhere. The tempered target in Eq.~\eqref{eq:target} permits a separate prior weight. Its corresponding functional is
\begin{subequations}
\begin{align}
F_{\vct{s}}^{(\gamma)}(\rho)&=\int\rho(\vct{y})\log\rho(\vct{y})d\vct{y}-\gamma\E_\rho\log p_{\mathrm{pre}}(Y)+\alpha\E_\rho D(\vct{s},Y)-\beta\E_\rho r(Y),\\
&=\KL(\rho\Vert\pi_{\vct{s}})-\log Z_{\vct{s}}.
\end{align}
\end{subequations}
This identity requires finite terms and a finite positive $Z_{\vct{s}}$; it does not require the prior power to be separately normalized.

\subsection{Conditional regularization versus marginal matching}
\label{app:conditional-information}
For sources $S\sim\mu$, let $T(\vct{y}\mid\vct{s})$ be the output density, $j_T(d\vct{s},d\vct{y})=\mu(d\vct{s})T(\vct{y}\mid\vct{s})d\vct{y}$, and $p_T$ its output marginal. Under integrability, source-wise minimizers also minimize $\E_S F_S(T(\cdot\mid S))$. The identity
\begin{equation}
\E_S\KL(T(\cdot\mid S)\Vert p_{\mathrm{pre}})=I_{j_T}(S;Y)+\KL(p_T\Vert p_{\mathrm{pre}})
\label{eq:conditional-information}
\end{equation}
shows that conditional prior regularization includes both mutual information and marginal matching. The source-distance term supplies the explicit preservation preference, without imposing a hard per-sample distance bound.

\section{Endpoint Realization and Its Scope}
\label{app:theory}
\begin{proposition}[Endpoint realization]
\label{prop:endpoint}
Let $Q_\tau$ be the zero-drift Brownian path law from Eq.~\eqref{eq:brownian-reference}, $q_\tau$ its endpoint density, and $P_{\vct{s}}^{\vct{b}}$ the path law with drift $\vct{b}$. Suppose the residual endpoint density $\bar\pi_{\vct{s}}$ is positive, has finite $\KL(\bar\pi_{\vct{s}}\Vert q_\tau)$, and its tilted path law
\begin{equation}
 \frac{dP_{\vct{s}}^\star}{dQ_\tau}
 =\frac{\bar\pi_{\vct{s}}(\vct{z}_1)}{q_\tau(\vct{z}_1)}
 \label{eq:optimal-path}
\end{equation}
is realizable by an admissible control. Under the Girsanov and integrability conditions in Appendix~\ref{app:theory-assumptions},
\begin{equation}
 J_{\vct{s}}(\vct{b})=\KL(P_{\vct{s}}^{\vct{b}}\Vert P_{\vct{s}}^\star).
 \label{eq:kl-identity}
\end{equation}
The unrestricted optimum therefore satisfies $\vct{s}+\vct{z}_1\sim\pi_{\vct{s}}$.
\end{proposition}

\subsection{Assumptions and proof of Proposition~\ref{prop:endpoint}}
\label{app:theory-assumptions}
Work on continuous residual paths starting at zero. Assume that $\bar\pi_{\vct{s}}$ is positive and normalized, with finite $\KL(\bar\pi_{\vct{s}}\Vert q_\tau)$. For each candidate control, assume the Girsanov stochastic exponential is a true martingale, and the relevant energies and logarithmic expectations are finite. Attaining the optimum within a controlled SDE also requires realizability of the tilted law. Smoothness of a target alone does not establish these conditions.

Set $w_{\vct{s}}(\vct{z})=\bar\pi_{\vct{s}}(\vct{z})/q_\tau(\vct{z})$. Its expectation is
\begin{equation}
 \E_{Q_\tau}[w_{\vct{s}}(\vct{z}_1)]
 =\int q_\tau(\vct{z})\frac{\bar\pi_{\vct{s}}(\vct{z})}{q_\tau(\vct{z})}d\vct{z}=1.
\end{equation}
Consequently $dP_{\vct{s}}^\star/dQ_\tau=w_{\vct{s}}(\vct{z}_1)=e^{-g_{\vct{s}}(\vct{z}_1)}$ defines a probability measure. For any measurable endpoint set $A$,
\begin{equation}
 P_{\vct{s}}^\star(\vct{z}_1\in A)
 =\int_Aq_\tau(\vct{z})w_{\vct{s}}(\vct{z})d\vct{z}
 =\int_A\bar\pi_{\vct{s}}(\vct{z})d\vct{z}.
\end{equation}
The tilt depends only on the endpoint, so the conditional path remains the reference bridge. Expanding the path KL gives
\begin{subequations}
\begin{align}
 \KL(P_{\vct{s}}^{\vct{b}}\Vert P_{\vct{s}}^\star)
 &=\E_{P_{\vct{s}}^{\vct{b}}}\left[
 \log\frac{dP_{\vct{s}}^{\vct{b}}}{dQ_\tau}
 -\log\frac{dP_{\vct{s}}^\star}{dQ_\tau}\right]\\
 &=\KL(P_{\vct{s}}^{\vct{b}}\Vert Q_\tau)+\E_{P_{\vct{s}}^{\vct{b}}}g_{\vct{s}}(\vct{z}_1)\\
 &=\E_{P_{\vct{s}}^{\vct{b}}}\left[
 \int_0^1\frac{\norm{\vct{b}(\vct{z}_\lambda,\lambda;\vct{s})}^2}{2\tau^2}d\lambda
 +g_{\vct{s}}(\vct{z}_1)\right].
\end{align}
\end{subequations}
The last equality uses Girsanov's path-energy identity. KL is minimized at the target path law whenever that law is realizable. Translation by $\vct{s}$ has unit Jacobian, giving output density $\pi_{\vct{s}}$.

We used normalized densities in $g_{\vct{s}}$. With the unnormalized energy
$F_{\vct{s}}=\alpha D(\vct{s},\cdot)-\gamma\log p_{\mathrm{pre}}-\beta r$, implementation may use $g^0_{\vct{s}}=\log q_\tau+F_{\vct{s}}(\vct{s}+\cdot)$. Then $g^0_{\vct{s}}=g_{\vct{s}}-\log\mathcal Z_{\vct{s}}$ and $J^0=\KL(P^{\vct{b}}_{\vct{s}}\Vert P^\star_{\vct{s}})-\log\mathcal Z_{\vct{s}}$; gradients and minimizers are unchanged.

\subsection{The realizing drift}
Assume
\begin{equation}
 h_{\vct{s}}(\vct{z},\lambda)=\E_{Q_\tau}[e^{-g_{\vct{s}}(\vct{z}_1)}\mid\vct{z}_\lambda=\vct{z}]
 \label{eq:app-h}
\end{equation}
is finite, positive and $C^{1,2}$ at interior times, and its logarithmic gradient defines an admissible nonexplosive diffusion. It solves the backward heat equation. It\^o's formula under $Q_\tau$ gives
\begin{equation}
 dh_{\vct{s}}(\vct{z}_\lambda,\lambda)=\tau\nabla h_{\vct{s}}(\vct{z}_\lambda,\lambda)^\top d\vct{w}_\lambda.
\end{equation}
The normalized process is the conditional likelihood-ratio martingale for the endpoint tilt. Girsanov therefore adds drift
\begin{equation}
 \vct{b}^\star(\vct{z},\lambda;\vct{s})=\tau^2\nabla\log h_{\vct{s}}(\vct{z},\lambda).
 \label{eq:app-doob}
\end{equation}
This realizes the desired path measure under the assumptions. A finite neural family need not contain this drift.

\subsection{Reference scale and source dependence}
For a fixed target, $q_\tau e^{-g_{\vct{s}}}=\bar\pi_{\vct{s}}$, so changing $\tau$ does not change the ideal endpoint. It does change the optimal drift, bridge geometry and numerical behavior. The source-distance penalty and conditional prior regularization have distinct roles, as shown by the mutual-information decomposition in Sec.~\ref{app:conditional-information}.

\section{Adjoint Sampling and Endpoint Regression}
\label{app:solver}
\begin{proposition}[Consistency of endpoint regression]
\label{prop:regression}
Consider the population counterpart of Eq.~\eqref{eq:ram-loss}: draw $\vct{z}_1$ from $\bar\pi_{\vct{s}}$, draw $\vct{z}_\lambda$ from its Brownian bridge, and replace the cached approximate label by the exact clean-score label $\vct{b}^{\mathrm{clean}}_{\vct{s}}(\vct{z}_1)$ in Eq.~\eqref{eq:drift-target}. Assume the labels have finite second moment and the target satisfies the regularity and tail conditions used below. With
$h_{\vct{s}}(\vct{z},\lambda)=\E_{Q_\tau}[e^{-g_{\vct{s}}(\vct{z}_1)}\mid\vct{z}_\lambda=\vct{z}]$,
the unrestricted population minimizer of this squared-error objective is
\begin{equation}
 \vct{b}^\star(\vct{z},\lambda;\vct{s})
 =\tau^2\nabla\log h_{\vct{s}}(\vct{z},\lambda)
 =\E_{P_{\vct{s}}^\star}[\vct{b}^{\mathrm{clean}}_{\vct{s}}(\vct{z}_1)\mid\vct{z}_\lambda=\vct{z}].
 \label{eq:regression-consistency}
\end{equation}
\end{proposition}

\subsection{Reciprocal projection and path KL}
For a path law $P$ with endpoint marginal $P_1$, define
\begin{equation}
 \mathcal R_{Q_\tau}(P)(d\omega)
 =\int Q_\tau(d\omega\mid Z_1=\vct{z})\,P_1(d\vct{z}),
 \label{eq:reciprocal-projection}
\end{equation}
where $\omega$ is a residual path starting at zero.

Let $P$ and $Q_\tau$ share the deterministic initial state, with $P\ll Q_\tau$ and finite path KL. Disintegrating over the terminal state gives
\begin{equation}
 \KL(P\Vert Q_\tau)=\KL(P_1\Vert q_\tau)
 +\E_{z\sim P_1}\KL(P(\cdot\mid Z_1=z)\Vert Q_\tau(\cdot\mid Z_1=z)).
\end{equation}
For $\mathcal R_{Q_\tau}(P)$ the second term vanishes and the first is unchanged. Terminal expectations are also unchanged, so the path-space objective $\KL(P\Vert Q_\tau)+\E_P g_s(Z_1)$ cannot increase. Interpreting the projected law as an admissible drift process additionally requires realizability and the usual Girsanov conditions. This is a property of the exact projection, not a monotonicity guarantee for finite neural regression steps.

\subsection{Control scaling and reference bridges}
Adjoint Matching formulates control learning as squared-error regression onto a lean-adjoint target~\citep{domingoenrich2025adjoint}. Using the normalized-control convention of Adjoint Sampling~\citep{havens2025adjointsampling}, let $\vct{u}=\vct{b}/\tau$. The physical drift is $\tau\vct{u}$ and the energy is $\norm{\vct{u}}^2/2$. Since the reference drift is zero and there is no running state cost, the lean reference adjoint has $\widetilde{\vct{a}}_\lambda=\nabla g_{\vct{s}}(\vct{z}_1)$ by Eq.~\eqref{eq:prelim-lean-adjoint}. The corresponding control-matching residual is
\begin{equation}
 \frac12\norm{\vct{u}_\theta+\tau\widetilde{\vct{a}}_\lambda}^2
 =\frac1{2\tau^2}\norm{\vct{b}_\theta+\tau^2\nabla g_{\vct{s}}(\vct{z}_1)}^2.
\end{equation}
Reciprocal Adjoint Matching replaces controlled intermediate states by reference-bridge samples conditional on endpoints from the current controller~\citep{havens2025adjointsampling}. Together with the physical-drift residual above, this gives Eq.~\eqref{eq:ram-loss} when endpoint gradients are cached and the ideal score is replaced by its estimator. The factor $1/(2\tau^2)$ is fixed and positive, so we omit it in the main-text MSE; the unrestricted minimizer is unchanged. Dividing by the latent dimension $d$ would likewise change only the overall loss scale. The adjoint refers to the zero-drift reference dynamics; it does not assert that the nonlinear learned controller has a constant exact adjoint. Expanding $g_{\vct{s}}=\log q_\tau-\log\bar\pi_{\vct{s}}$ gives the exact clean-score label before introducing a practical score estimator:
\begin{subequations}
\label{eq:drift-target}
\begin{align}
 \vct{b}^{\mathrm{clean}}_{\vct{s}}(\vct{z}_1)
 &:=-\tau^2\nabla_{\vct{z}_1}g_{\vct{s}}(\vct{z}_1)\\
 &=\vct{z}_1+\tau^2\!\left[
 \gamma\nabla_{\vct{y}}\log p_{\mathrm{pre}}(\vct{y})
 -\alpha\nabla_{\vct{y}}D(\vct{s},\vct{y})
 +\beta\nabla_{\vct{y}}r(\vct{y})\right]_{\vct{y}=\vct{s}+\vct{z}_1}.
 \end{align}
\end{subequations}
The $\vct{z}_1$ term comes from $-\tau^2\nabla\log q_\tau(\vct{z}_1)$; inserting the quadratic preservation cost gives the clean analogue of Eq.~\eqref{eq:quadratic-label}. Replacing the clean score by $\widehat{\vct{s}}_{\mathrm{pre}}$ yields the implemented label in that equation.

Under $Q_\tau$, $\Cov(\vct{z}_\lambda,\vct{z}_1)=\tau^2\lambda\mat I$, $\Var(\vct{z}_\lambda)=\tau^2\lambda\mat I$ and $\Var(\vct{z}_1)=\tau^2\mat I$. Gaussian conditioning gives mean $\lambda\vct{z}_1$ and covariance
$\tau^2\lambda\mat I-(\tau^2\lambda\mat I)(\tau^2\mat I)^{-1}(\tau^2\lambda\mat I)=\tau^2\lambda(1-\lambda)\mat I$, proving Eq.~\eqref{eq:brownian-reference}.

\subsection{Proof of Proposition~\ref{prop:regression}}
Let $k(\vct{v}-\vct{z})$ denote the Gaussian transition density with covariance $\tau^2(1-\lambda)\mat I$. Then
\begin{equation}
 h_{\vct{s}}(\vct{z},\lambda)=\int k(\vct{v}-\vct{z})e^{-g_{\vct{s}}(\vct{v})}d\vct{v}.
\end{equation}
Assume differentiation may pass under the integral and boundary terms vanish. Using $\nabla_{\vct{z}}k=-\nabla_{\vct{v}}k$ and integrating by parts,
\begin{subequations}
\begin{align}
 \nabla_{\vct{z}}h_{\vct{s}}
 &=-\int\nabla_{\vct{v}}k(\vct{v}-\vct{z})e^{-g_{\vct{s}}(\vct{v})}d\vct{v}\\
 &=\int k(\vct{v}-\vct{z})\nabla_{\vct{v}}e^{-g_{\vct{s}}(\vct{v})}d\vct{v}\\
 &=-\int k(\vct{v}-\vct{z})e^{-g_{\vct{s}}(\vct{v})}\nabla g_{\vct{s}}(\vct{v})d\vct{v}.
\end{align}
\end{subequations}
The conditional endpoint density under $P_{\vct{s}}^\star$ is $k(\vct{v}-\vct{z})e^{-g_{\vct{s}}(\vct{v})}/h_{\vct{s}}(\vct{z},\lambda)$. Dividing by $h_{\vct{s}}$ and multiplying by $\tau^2$ gives Eq.~\eqref{eq:regression-consistency}.

The endpoint tilt leaves reference conditional bridges unchanged; hence sampling an endpoint from $\bar\pi_{\vct{s}}$ and a reference bridge produces the joint marginal of $P_{\vct{s}}^\star$. For the square-integrable ideal label $\vct{b}^{\mathrm{clean}}_{\vct{s}}(\vct{z}_1)$, let $\vct{m}(\vct{z}_\lambda)=\E_{P_{\vct{s}}^\star}[\vct{b}^{\mathrm{clean}}_{\vct{s}}(\vct{z}_1)\mid\vct{z}_\lambda]$. Conditional-variance decomposition writes the expected squared error as $\norm{\vct{b}(\vct{z}_\lambda,\lambda;\vct{s})-\vct{m}(\vct{z}_\lambda)}^2$ plus a term independent of the predictor; the cross term vanishes by the definition of $\vct{m}$. The optimal predictor is therefore the conditional mean derived above.

\subsection{Finite updates and replay}
For a general frozen endpoint law $\rho_k$, the regression mean is taken under $\rho_k(\vct{z}_1)Q_\tau(d\vct{z}_\lambda\mid\vct{z}_1)$. The proof establishes the target fixed point, not a one-step exact solution or global convergence. In practice, endpoints follow the current controller law and Eq.~\eqref{eq:quadratic-label} uses an approximate prior score; either change breaks the exact fixed-point premise. Equation~\eqref{eq:ram-loss} conditions on one detached endpoint and label per refresh. This empirical objective need not have the population minimizer even if that endpoint came from the target. Resampling bridges supplies stochastic gradients of this frozen empirical objective, not new independent terminal information. The next refresh replaces the stored pair; retaining older endpoints would instead define a replay variant.

\section{Endpoint Prior and Reward Oracles}
\label{app:oracle}

The ideal theory uses the clean prior. A conservative surrogate replaces its energy $-\log p_{\mathrm{pre}}$ in Eq.~\eqref{eq:target}. An exact clean score, a noisy diffusion score, and an arbitrary guidance vector are not interchangeable. One possible explicit surrogate is as follows. Let $\ell$ be a noise level sampled from a fixed distribution, $a_\ell$ and $\nu_\ell>0$ fixed corruption coefficients, $p_\ell^{\mathrm{pre}}$ the corresponding pretrained noisy density, and $\vct{\epsilon}$ standard Gaussian noise. Define
\begin{equation}
 \mathcal P_{\mathrm{noise}}(\vct{y})=
 \E_{\ell,\vct{\epsilon}}
 [-\log p_\ell^{\mathrm{pre}}(a_\ell\vct{y}+\nu_\ell\vct{\epsilon})].
 \label{eq:app-prior}
\end{equation}
If differentiation and expectation may be exchanged, its gradient is
$-\E_{\ell,\vct{\epsilon}}[a_\ell\nabla\log p_\ell^{\mathrm{pre}}
 (a_\ell\vct{y}+\nu_\ell\vct{\epsilon})]$.
A pretrained score predictor can supply a Monte Carlo estimate by forward evaluation. This avoids differentiating the score network, but targets $\mathcal P_{\mathrm{noise}}$, not necessarily $-\log p_{\mathrm{pre}}$.
For example, when the score predictor returns $\nabla\log p_\ell^{\mathrm{pre}}$ at the corrupted input, a single draw gives
\begin{equation}
 \widehat{\vct{s}}_{\mathrm{pre}}(\vct{y};\ell,\vct{\epsilon})
 =a_\ell\nabla\log p_\ell^{\mathrm{pre}}
      (a_\ell\vct{y}+\nu_\ell\vct{\epsilon}),
 \qquad
 \E_{\ell,\vct{\epsilon}}[\widehat{\vct{s}}_{\mathrm{pre}}]
 =-\nabla\mathcal P_{\mathrm{noise}}(\vct{y}).
 \label{eq:app-score-estimate}
\end{equation}
If this estimator is exact in expectation and the normalizer exists, it is associated with the surrogate endpoint density
$\widetilde\pi_{\vct{s}}(\vct{y})\propto
\exp[-\gamma\mathcal P_{\mathrm{noise}}(\vct{y})
-\alpha D(\vct{s},\vct{y})+\beta r(\vct{y})]$.
The score estimator still varies between queries and the surrogate generally differs from the clean-prior target.
For the quadratic cost $D(\vct{s},\vct{y})=\norm{\vct{y}-\vct{s}}^2/2$, we have $\nabla_{\vct{y}}D=\vct{z}_1$. Together with $\nabla\log q_\tau(\vct{z}_1)=-\vct{z}_1/\tau^2$, Eq.~\eqref{eq:drift-target} reduces to Eq.~\eqref{eq:quadratic-label} after the clean prior score is replaced by $\widehat{\vct{s}}_{\mathrm{pre}}$. If the estimate is the Monte Carlo gradient of Eq.~\eqref{eq:app-prior}, its expectation is a gradient field, but it corresponds to a noisy-prior surrogate rather than the clean density in Eq.~\eqref{eq:target}. A single sampled estimate has variance; the other approximations in the score predictor may add bias.

The prior-query settings are given in Appendix~\ref{app:experiments}. Inexact score predictions need not recover the clean-prior target; Efficient Adjoint Matching also discusses the difficulty of obtaining endpoint scores~\citep{shin2026eam}.

For image-space rewards, gradients must pass from $R(\mathcal D(\vct{y}))$ back to $\vct{y}$ even though both networks' weights are frozen. Intermediate activation graphs are released after obtaining the detached endpoint gradient. Source latents and fixed text features can be cached. Learned controller features can only be cached across updates if their producing parameters do not change.

\section{Controller and Numerical Implementation}
\label{app:implementation}

The main-text parameterization in Eq.~\eqref{eq:parameterization} combines a trainable latent offset with a state- and time-dependent convolutional correction. We give its architecture and numerical implementation here.

For a latent tensor of shape $C\times H\times W$, the controller concatenates $\vct{z}$, $\vct{s}$ and a broadcast scalar time into $2C+1$ channels. The nonlinear branch consists of three $3\times3$ convolutions with padding one, channel widths $(2C+1,h,h,C)$, and SiLU after the first two layers. The translation contains $CHW$ trainable scalars. Including all biases, the total number of parameters is
\begin{equation}
 N_\theta=CHW+9h^2+27Ch+11h+C.
\end{equation}
For example, $C=4$, $H=W=64$ and $h=16$ give $20{,}596$ parameters including the translation. This is an architecture example, not a claim that every experimental setting uses those dimensions. The translation and final convolution start at zero; other layers use their usual initialization.

Euler--Maruyama simulates Eq.~\eqref{eq:faster-sde} with $\Delta \lambda=1/N$ and $N=16$ and independent $\vct{\xi}_n\sim\Normal(0,\mat I)$:
\begin{equation}
 \vct{z}_{n+1}=\vct{z}_n+\Delta \lambda\,\vct{b}_\theta(\vct{z}_n,n\Delta \lambda;\vct{s})
 +\tau\sqrt{\Delta \lambda}\,\vct{\xi}_n.
\end{equation} Endpoint simulation is detached from controller optimization. The prior model remains frozen and is queried outside the SDE steps. Reward-input gradients still pass through the frozen decoder and reward network; freezing weights does not eliminate those derivatives. After fitting, a fresh independent trajectory produces the output, rather than selecting the best training endpoint.

Experimental settings are listed in Appendix~\ref{app:experiments} and Table~\ref{tab:faster-hyperparameters}. Each refresh replaces the cached endpoint and label; older endpoints are not retained.

\subsection{Endpoint-query reuse}
With $K$ endpoint refreshes and one candidate per refresh, the basic loop computes $K$ endpoint targets, up to the chosen number of noisy-score samples per candidate. Its $KM$ controller updates reuse those targets and draw new bridge times and noises. Each refresh also requires $N$ evaluations of the small controller for Euler--Maruyama simulation. These counts describe the algorithmic placement of expensive queries; actual runtime depends on decoder, score-model, controller, and optimizer costs. Differentiation through the decoder is confined to the endpoint-query step. Since the targets are detached, the inner regression loop does not build or retain computation graphs for the pretrained model.

\section{Evaluation Protocol and Baseline Details}
\label{app:experiments}

\paragraph{Frozen inputs.}
We instantiate the four tasks of \citet{chang2026itoc} with 300 frozen source/condition pairs per task. Human-preference sources use prompts from the Reward Feedback Learning (ReFL) data released with ImageReward~\citep{xu2023imagereward}; style sources are from the Pick-a-Pic test split~\citep{kirstain2023pickapic} and are paired with five fixed style-reference images; counterfactual sources are ImageNet-1k images~\citep{deng2009imagenet}; facial-edit sources are CelebA-HQ images~\citep{karras2018progressive} with single-attribute text requests. All methods for a given model/task use the same source-image hashes, target text or class, and style reference when applicable. For counterfactual targets, we select a nearby ImageNet class using a WordNet IS-A graph~\citep{miller1995wordnet}.

\paragraph{Baseline implementations.}
\begin{itemize}
\item \textbf{ITOC}~\citep{chang2026itoc} iteratively optimizes an image-conditioned sampling trajectory using reward adjoints.
\item \textbf{Gradient Ascent (GA)} directly updates the editable latent using the task-reward gradient.
\item \textbf{DPS}~\citep{chung2023dps} applies posterior-style reward guidance during reverse sampling after source inversion. This is the source-editing adaptation used in the ITOC comparison suite.
\item \textbf{FreeDoM}~\citep{yu2023freedom} uses its training-free energy guidance after source inversion, with the same frozen source set.
\item \textbf{TFG}~\citep{ye2024tfg} uses training-free reward guidance during the corresponding inversion-conditioned reverse trajectory.
\item \textbf{MPGD}~\citep{he2024mpgd} updates a predicted clean endpoint with the reward gradient. On SD1.5 we use an MPGD-LDM source-edit variant; on SD3 we use a flow-edit adaptation.
\end{itemize}
The model, source set, task reward, and evaluator are paired within each comparison. Differences in inversion and guidance mechanisms are intrinsic to the baselines. We report mean metric values.

\paragraph{Target and validation definitions.}
Human preference directly optimizes ImageReward and additionally reports HPSv2, image--text CLIPScore~\citep{hessel2021clipscore}, computed as the cosine similarity of CLIP ViT-L/14 image and text embeddings, and Aesthetic. Style directly minimizes the released positive GramDiff and validates with CLIP-I similarity to the style reference and DINO ViT-S/16 similarity. Counterfactual editing optimizes the robust classifier's target-class logit and validates with CLIPScore for the target-class text. Facial editing optimizes CLIPScore for the attribute request and validates with ImageReward and HPSv2. Source preservation is additionally available as LPIPS~\citep{zhang2018lpips} and source--output CLIP-I in the per-image evaluator records; these auxiliary columns are kept separate from the target/validation tables.

The reward implementations use ImageReward-v1.0~\citep{xu2023imagereward}, OpenAI CLIP ViT-L/14 for image--text cosine similarity~\citep{radford2021clip}, and the robust ResNet-50 checkpoint \texttt{imagenet\_l2\_3\_0.pt} distributed with the robustness library~\citep{engstrom2019robustness}. The counterfactual reward is the target logit, not a target-text score. Style GramDiff uses non-class tokens from \texttt{hidden\_states[2]} of \texttt{openai/clip-vit-base-patch16}; the optimization reward is the negative Frobenius difference between the output and reference Gram matrices, scaled by $1/100$. The table reports the corresponding positive difference. DINO validation uses ViT-S/16~\citep{caron2021dino}; HPSv2 follows \citet{wu2023hpsv2}. The Aesthetic evaluator uses the released CLIP-feature predictor from \url{https://github.com/christophschuhmann/improved-aesthetic-predictor}.

\paragraph{Paired comparison and selection.}
Every model--task comparison uses the same 300 source--condition pairs for all seven methods. FASTER parameters were selected using earlier pilot evaluations, before the full-set evaluation. In the tables, bold values indicate the best result for each metric. All required edited-image metrics are finite.

\paragraph{FASTER configurations.}
The selected runs use $K=20$ outer refreshes and $M=2$ controller updates per refresh. The latent SDE uses 16 steps for both SD1.5 and SD3, at working resolutions of $512^2$ and $768^2$, respectively. Each refresh samples one endpoint; each controller update independently draws four control times $\lambda$ and four Gaussian noises conditioned on that endpoint, then averages their regression losses for one optimizer step. The controller has 16 hidden channels, a trainable terminal translation, and a nonlinear branch. We fit it with mean-squared error and Adam, using learning rates $0.02$ for the translation and $0.001$ for the branch. Bridge times are uniform on $[0,1]$; the seed is zero. Each enabled prior query uses one sample with antithetic sampling enabled. The prior-score query window is diffusion steps $[100,400]$ for SD1.5 and flow noise times $[0.1,0.4]$ for SD3. The SD3 style objective uses symmetric bicubic preprocessing; the SD1.5 style objective uses the released preprocessing. The quantitative benchmark runs do not use the optional Editing Term; its separate noise-time windows are reported in Appendix~\ref{app:edit-term}. For every row below, $\alpha=(\alpha\tau^2)/\tau^2$, $\beta=c_r/\tau^2$, and $\gamma=c_p/\tau^2$.
\begin{table}[H]
\caption{Selected FASTER settings for the quantitative benchmarks.}
\label{tab:faster-hyperparameters}
\centering
\small
\begin{tabular}{llrrrr}
\toprule
Backbone & Task & $\alpha\tau^2$ & $\tau$ & $c_r$ & $c_p$ \\
\midrule
SD1.5 & Human & 1 & 0.0125 & 320 & 0.1 \\
SD1.5 & Style & 0.01 & 0.002 & 10240 & 0 \\
SD1.5 & Counterfactual & 1 & 0.002 & 2560 & 0 \\
SD1.5 & Facial edit & 1 & 0.00625 & 1280 & 0.01 \\
SD3 & Human & 1 & 0.0125 & 320 & 0 \\
SD3 & Style & 0.01 & 0.0125 & 5120 & 0 \\
SD3 & Counterfactual & 1 & 0.0125 & 20480 & 0 \\
SD3 & Facial edit & 1 & 0.0125 & 20480 & 0.1 \\
\bottomrule
\end{tabular}
\end{table}

\paragraph{Baseline configurations.}
Table~\ref{tab:baseline-hyperparameters} lists the task-dependent settings. For ITOC, $(n,w,\ell)$ denotes trajectory-optimization iterations, reward weight, and learning rate; inversion depth is $0.5$ on SD1.5 and $0.7$ on SD3. GA makes 100 gradient-ascent steps in latent space with the listed learning rate. DPS, FreeDoM, and TFG use inversion depth $0.7$ and the listed reward-guidance base coefficient $\rho$. DPS makes one guided reverse pass; FreeDoM uses two recurrence steps; TFG uses one recurrence with four inner gradient steps, the listed base coefficient $\mu$, and a constant noise parameter $\sigma=0.1$. DPS, FreeDoM, and TFG all use the increasing $\rho$ schedule; DPS and FreeDoM set $\mu=0$ and disable the optional guidance-noise schedule. TFG uses an increasing $\mu$ schedule, a constant-noise schedule with $\sigma=0.1$, one noise draw per step, and unit gradient clipping scale. MPGD uses the listed clean-endpoint gradient step $\lambda$ at every reverse step, with depth $0.7$, constant guidance throughout the reverse interval, and no text classifier-free guidance. Its SD1.5 implementation uses the paper-style DDIM update; its SD3 implementation uses the flow adaptation described above. All baselines use 50 scheduler steps on SD1.5 or 28 on SD3, working resolutions of $512^2$ or $768^2$, and seed zero. All baselines use the same frozen source--condition pairs within each model and task.
\begin{table}[H]
\caption{Task-specific baseline settings used for the reported results. The notation and shared settings are defined in the text.}
\label{tab:baseline-hyperparameters}
\centering
\setlength{\tabcolsep}{3pt}
\resizebox{\linewidth}{!}{%
\begin{tabular}{llcccccc}
\toprule
Model & Task & ITOC $(n,w,\ell)$ & GA $\ell$ & DPS $\rho$ & FreeDoM $\rho$ & TFG $(\rho,\mu)$ & MPGD $\lambda$ \\
\midrule
SD3 & Human & $(15,500,0.005)$ & 2 & 5 & 5 & $(5,1)$ & 1 \\
SD3 & Style & $(15,1000,0.005)$ & 3 & 50 & 20 & $(40,2.5)$ & 12 \\
SD3 & Counterfactual & $(15,200,0.005)$ & 1 & 7 & 3 & $(7,0.5)$ & 7 \\
SD3 & Face & $(15,1000,0.005)$ & 1.5 & 75 & 30 & $(60,2.5)$ & 10 \\
\midrule[.8pt]
SD1.5 & Human & $(20,500,0.005)$ & 2 & 3 & 1 & $(1,0.5)$ & 1 \\
SD1.5 & Style & $(20,200,0.01)$ & 3 & 15 & 7.5 & $(10,1)$ & 12 \\
SD1.5 & Counterfactual & $(20,50,0.005)$ & 1 & 1 & 0.4 & $(1,0.1)$ & 7 \\
SD1.5 & Face & $(20,1000,0.005)$ & 1.5 & 40 & 20 & $(30,2.5)$ & 10 \\
\bottomrule
\end{tabular}}
\end{table}

\paragraph{SD1.5 results.}
Table~\ref{tab:app-sd15} reports the full SD1.5 comparison. FASTER leads the human-preference CLIPScore and Aesthetic validation columns, while its ImageReward is within 1.2\% of the best GA result. It leads both the optimized CLIPScore and ImageReward in facial editing. Style transfer attains the lowest GramDiff but trails the best style validation scores, especially DINO. In counterfactual editing its CLIPScore is close to MPGD, yet its target logit remains lower.
\begin{table}[!ht]
\caption{Results on Stable Diffusion 1.5. Each task includes a target metric, which is directly optimized, and validation metrics used to assess potential reward hacking. Arrows indicate whether higher ($\uparrow$) or lower ($\downarrow$) values are better. Bold values indicate the best result for each metric.}
\label{tab:app-sd15}
\centering
\setlength{\tabcolsep}{3pt}
\resizebox{\linewidth}{!}{%
\begin{tabular}{lrrrrrrr}
\toprule
Metric & ITOC & GA & DPS & FreeDoM & TFG & MPGD & \textbf{FASTER} \\
\midrule
\rowcolor{taskband}
\multicolumn{8}{l}{\textbf{Human preference}} \\
{\scriptsize\color{brightorange}TARGET}\,\textbf{ImageReward}$\uparrow$ & 1.8981 & \textbf{1.9532} & 1.5374 & 1.3118 & 1.6908 & 1.1938 & 1.9316 \\
\hdashline[5pt/3pt]\noalign{\vskip 2pt}
{\scriptsize\color{softpurple}VALIDATION}\,HPSv2$\uparrow$ & \textbf{0.2666} & 0.2615 & 0.2578 & 0.2555 & 0.2606 & 0.2535 & 0.2649 \\
{\scriptsize\color{softpurple}VALIDATION}\,CLIPScore$\uparrow$ & 0.2773 & 0.2743 & 0.2546 & 0.2366 & 0.2642 & 0.2531 & \textbf{0.2798} \\
{\scriptsize\color{softpurple}VALIDATION}\,Aesthetic$\uparrow$ & 5.8058 & 5.5443 & 5.4939 & 5.4869 & 5.5269 & 5.2346 & \textbf{5.8075} \\
\midrule[.8pt]
\rowcolor{taskband}
\multicolumn{8}{l}{\textbf{Style transfer}} \\
{\scriptsize\color{brightorange}TARGET}\,\textbf{GramDiff}$\downarrow$ & 9.0027 & 7.7940 & 5.8574 & 5.4821 & 6.0595 & 6.9080 & \textbf{5.0592} \\
\hdashline[5pt/3pt]\noalign{\vskip 2pt}
{\scriptsize\color{softpurple}VALIDATION}\,CLIP-I$_{\mathrm{sty}}$$\uparrow$ & 0.4944 & 0.5045 & \textbf{0.5282} & 0.5280 & 0.5252 & 0.5233 & 0.5190 \\
{\scriptsize\color{softpurple}VALIDATION}\,DINO$_{\mathrm{sty}}$$\uparrow$ & 0.1585 & 0.1491 & 0.2471 & \textbf{0.2645} & 0.2460 & 0.1875 & 0.1895 \\
\midrule[.8pt]
\rowcolor{taskband}
\multicolumn{8}{l}{\textbf{Counterfactual generation}} \\
{\scriptsize\color{brightorange}TARGET}\,\textbf{Logit$_{\mathrm{tgt}}$}$\uparrow$ & 19.7998 & 27.5803 & 17.0912 & 14.9241 & 18.9315 & \textbf{31.7846} & 28.8259 \\
\hdashline[5pt/3pt]\noalign{\vskip 2pt}
{\scriptsize\color{softpurple}VALIDATION}\,CLIPScore$\uparrow$ & 0.1803 & 0.2017 & 0.1800 & 0.1702 & 0.1842 & \textbf{0.2190} & 0.2184 \\
\midrule[.8pt]
\rowcolor{taskband}
\multicolumn{8}{l}{\textbf{Text-guided facial editing}} \\
{\scriptsize\color{brightorange}TARGET}\,\textbf{CLIPScore}$\uparrow$ & 0.2886 & 0.2544 & 0.2763 & 0.2716 & 0.2831 & 0.2554 & \textbf{0.3266} \\
\hdashline[5pt/3pt]\noalign{\vskip 2pt}
{\scriptsize\color{softpurple}VALIDATION}\,ImageReward$\uparrow$ & -0.6209 & -0.7547 & -0.6610 & -0.5575 & -0.5089 & -0.7707 & \textbf{-0.4492} \\
{\scriptsize\color{softpurple}VALIDATION}\,HPSv2$\uparrow$ & 0.2559 & \textbf{0.2565} & 0.2511 & 0.2536 & 0.2521 & 0.2509 & 0.2542 \\
\bottomrule
\end{tabular}}
\end{table}

\begin{figure}[H]
\centering
\includegraphics[width=\linewidth]{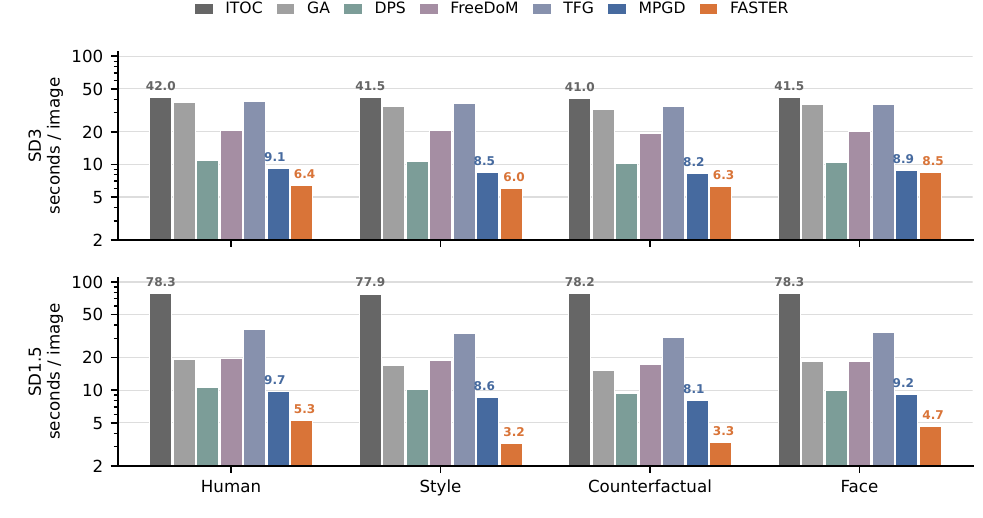}
\caption{Editing time on one H20 across four tasks and two model families, averaged over three images per setting. The data and timing scope match Fig.~\ref{fig:itoc-faster-runtime}; the vertical axis is logarithmic.}
\label{fig:four-task-runtime}
\end{figure}

\paragraph{Editing time and memory.}
Figure~\ref{fig:four-task-runtime} extends Fig.~\ref{fig:itoc-faster-runtime} to all seven methods. All methods use the same timing scope: the GPU editing computation, including source inversion where required and all decoder and reward evaluations needed for optimization. We exclude model and image loading, input preprocessing and source encoding, final output decoding and saving, and offline metric evaluation. Timings are synchronized elapsed times for the editing routine, including its execution overhead rather than a sum of GPU-kernel durations. After excluding each worker's first image, we average the same three source images per task on one H20; speedups use unrounded means.

A separate $768\times768$ SD3 ImageReward edit with $K=20$ on one RTX 4090 D takes 7.72\,s and peaks at 18.75\,GiB of live GPU tensor allocation. The text encoders are moved to CPU after prompt encoding, while the transformer, VAE, and ImageReward remain on the GPU. Peak memory is measured with PyTorch's peak allocated-memory counter, excluding allocator cache; time follows the editing-only scope above.

\begin{figure}[H]
\centering
\includegraphics[width=\linewidth]{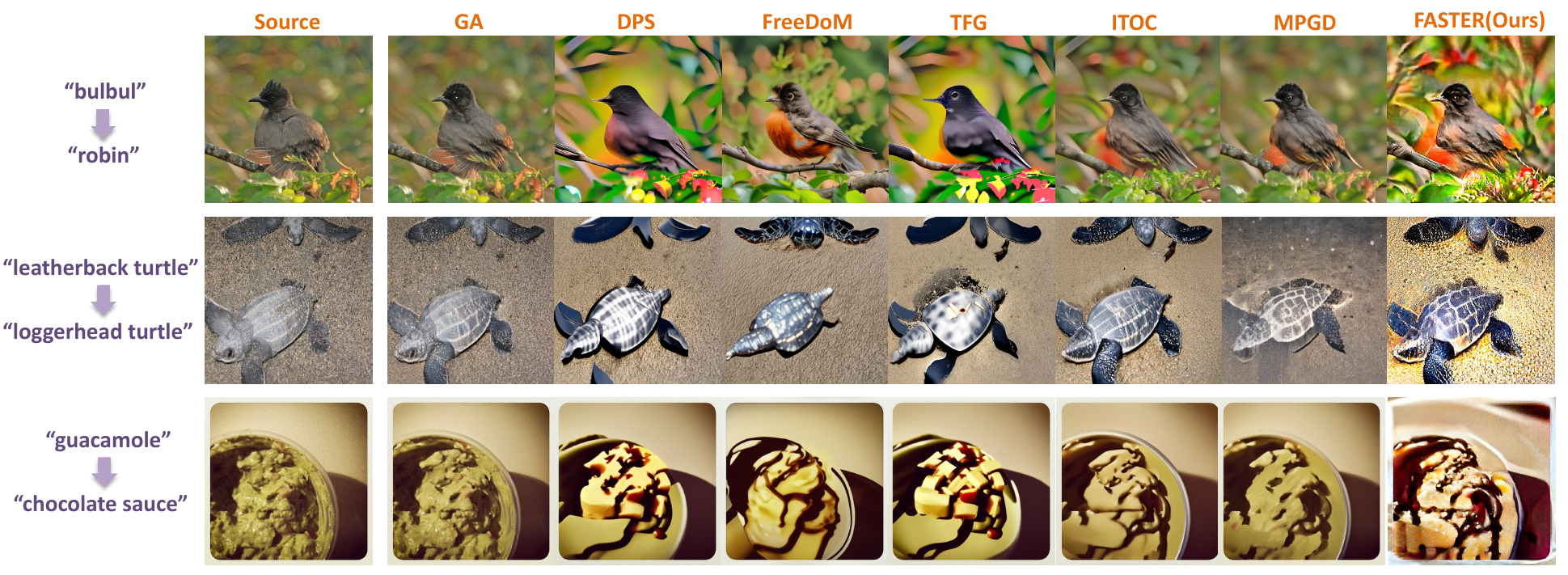}
\caption{Additional matched counterfactual edits. Source and target classes are shown at left; method columns follow Fig.~\ref{fig:sd3-qualitative}.}
\label{fig:app-counterfactual-examples}
\end{figure}

\begin{figure}[H]
\centering
\includegraphics[width=\linewidth]{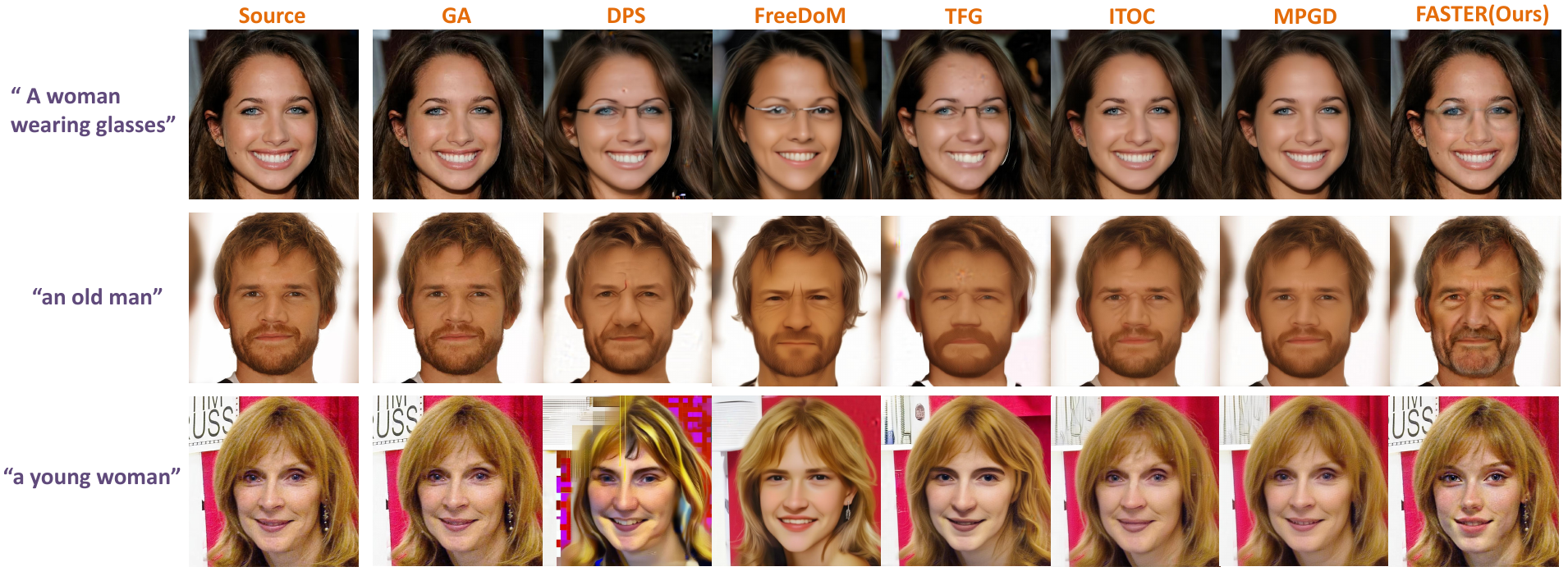}
\caption{Additional matched facial edits. The requested text prompts are shown at left; method columns follow Fig.~\ref{fig:sd3-qualitative}.}
\label{fig:app-face-examples}
\end{figure}

\begin{figure}[H]
\centering
\includegraphics[width=\linewidth]{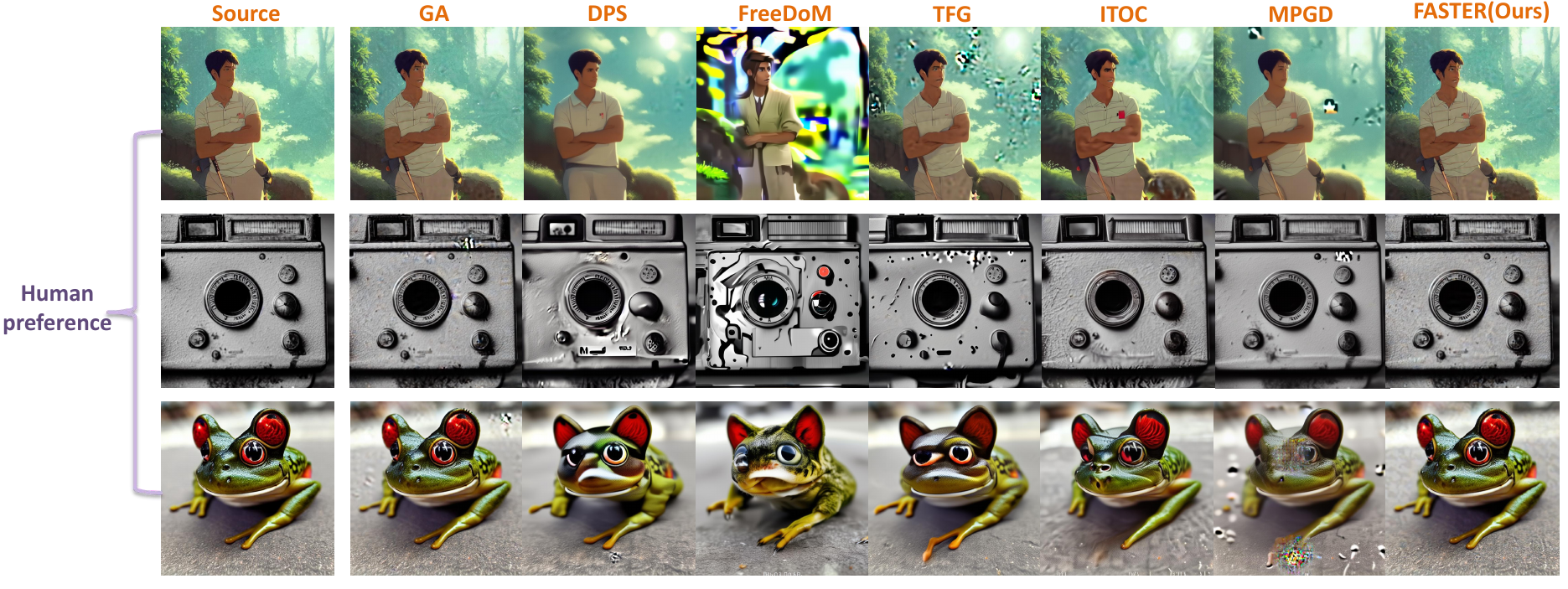}
\caption{Additional matched human-preference examples, with the source image and outputs from each method.}
\label{fig:app-human-examples}
\end{figure}

\begin{figure}[H]
\centering
\includegraphics[width=\linewidth,trim=86bp 178bp 99bp 60bp,clip]{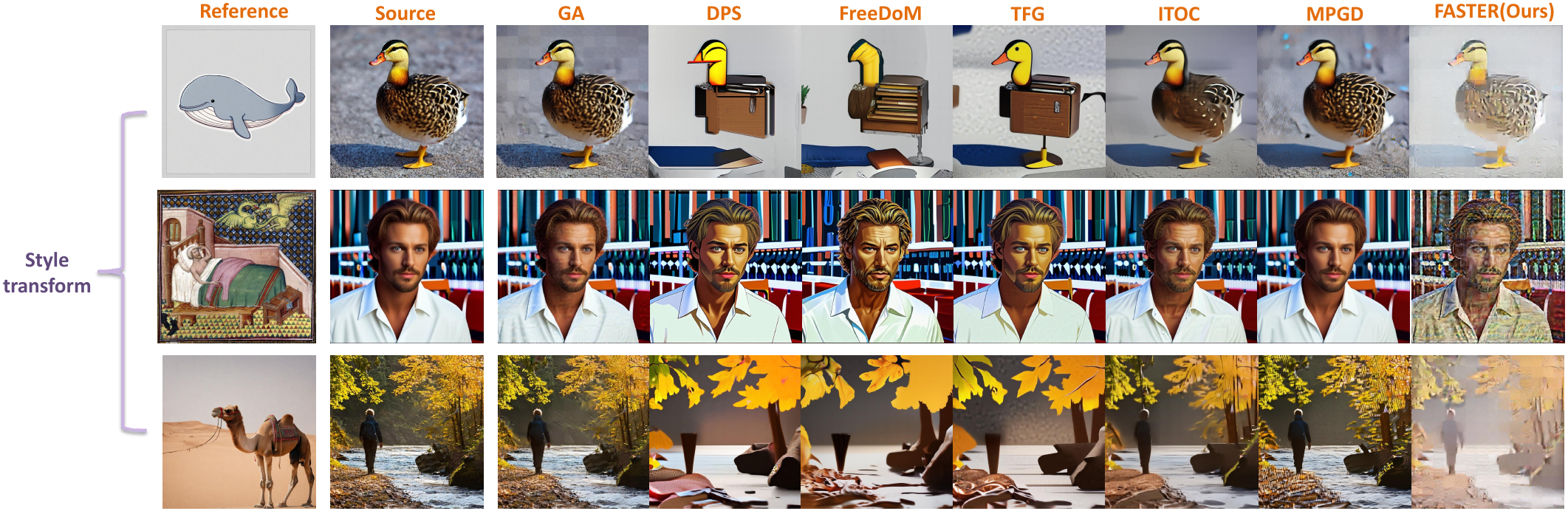}
\caption{Additional style-transfer examples. The style reference and source image occupy the first two columns; method columns follow Fig.~\ref{fig:sd3-qualitative}.}
\label{fig:app-style-examples}
\end{figure}

\paragraph{Additional visual results.}
Figures~\ref{fig:app-counterfactual-examples}--\ref{fig:app-style-examples} show further counterfactual, facial, human-preference, and style-transfer edits, respectively.

\FloatBarrier

\section{Editing Term: Conditional-Score Difference}
\label{app:edit-term}

\paragraph{From velocity to an endpoint edit field.}
Use the source and target conditions $c_{\mathrm{src}}$ and $c_{\mathrm{tgt}}$ from Sec.~\ref{sec:method-edit}. For the SD3 flow convention, a clean latent $Y$ and Gaussian noise $\epsilon$ produce $x_t=(1-t)Y+t\epsilon$, and the exact conditional velocity is $v_c(x_t,t)=\mathbb E[\epsilon-Y\mid x_t,c]$. Rearranging the interpolation gives $\epsilon=x_t+(1-t)(\epsilon-Y)$. The Gaussian-mixture score identity therefore yields
\begin{equation}
 s_c(x_t,t):=\nabla_{x_t}\log p_t(x_t\mid c)
 =-\frac{\mathbb E[\epsilon\mid x_t,c]}{t}
 =-\frac{x_t+(1-t)v_c(x_t,t)}{t}.
 \label{eq:edit-score-from-velocity}
\end{equation}
For $t\sim\nu$ in a fixed interval $0<t_{\min}<t_{\max}<1$ and independent Gaussian noise, we use the target-to-source conditional log-density ratio as a smoothed edit potential at endpoint $y$:
\begin{equation}
 E_{\mathrm{edit}}(y)=\mathbb E_{t,\epsilon}\!\left[
 \log p_t((1-t)y+t\epsilon\mid c_{\mathrm{tgt}})
 -\log p_t((1-t)y+t\epsilon\mid c_{\mathrm{src}})\right].
 \label{eq:edit-potential}
\end{equation}
Assume the conditional densities are positive and differentiable, with an integrable gradient bound permitting differentiation under the expectation. For fixed $t$ and $\epsilon$, the Jacobian of $x_t$ with respect to $y$ is $(1-t)I$. Consequently, the pullback of the \emph{target-minus-source score} is
\begin{subequations}
\label{eq:edit-pullback}
\begin{align}
 \nabla_y E_{\mathrm{edit}}(y)
 &=\mathbb E_{t,\epsilon}\!\left[(1-t)
 (s_{c_{\mathrm{tgt}}}-s_{c_{\mathrm{src}}})(x_t,t)\right]\\
 &=\mathbb E_{t,\epsilon}\!\left[\frac{(1-t)^2}{t}
 (v_{c_{\mathrm{src}}}-v_{c_{\mathrm{tgt}}})(x_t,t)\right].
\end{align}
\end{subequations}
This proves Eq.~\eqref{eq:semantic-gradient}. The velocity order reverses because Eq.~\eqref{eq:edit-score-from-velocity} has a minus sign. This is analogous to a conditional-guidance residual, with the source prompt rather than an unconditional prompt as reference. The implementation estimates Eq.~\eqref{eq:edit-pullback} using paired source/target velocity evaluations at the same $x_t$ and antithetic noise, then adds $c_e\nabla_y E_{\mathrm{edit}}$ to the FASTER endpoint label. The pullback itself is linear and does not require backpropagation through the SD3 transformer. Equation~\eqref{eq:edit-pullback} is exact for exact conditional scores; learned velocities yield an approximate edit field in practice.

\begin{figure}[H]
\centering
\includegraphics[width=\linewidth,trim=23bp 39bp 22bp 21bp,clip]{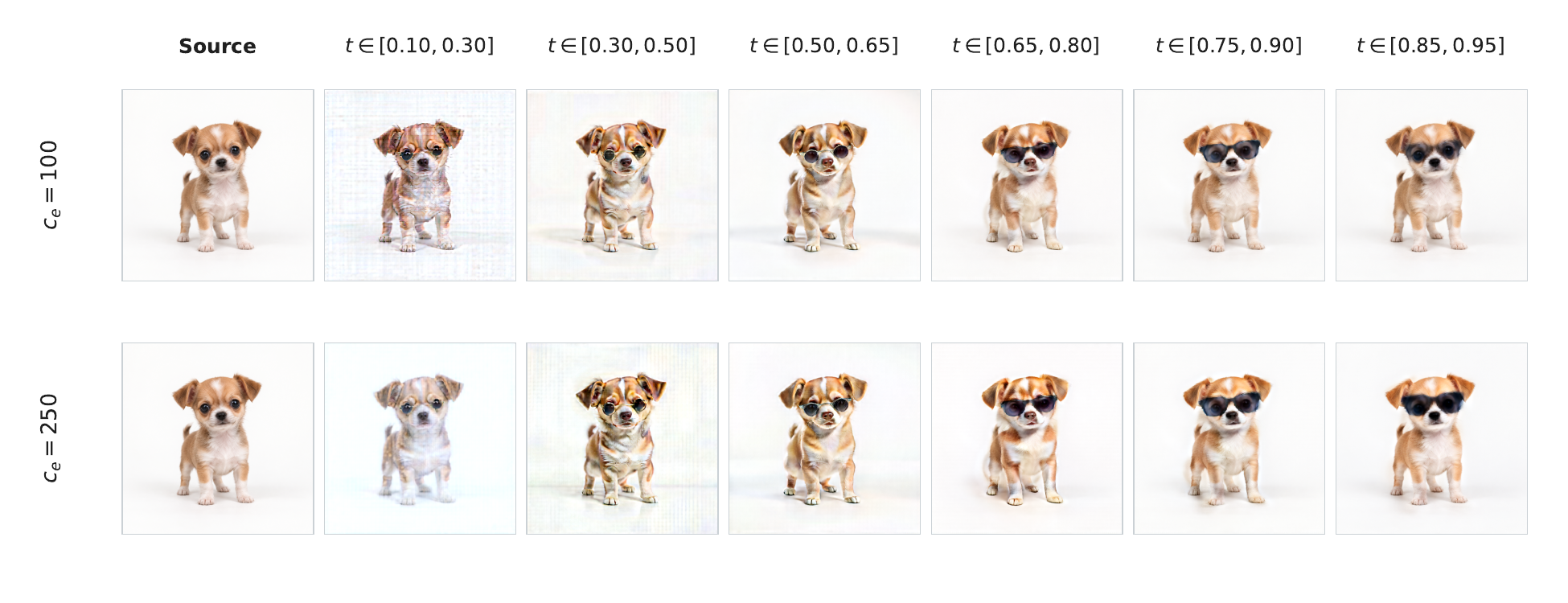}
\caption{Adding sunglasses to the same puppy with the Editing Term. Columns vary the noise-time interval and rows vary $c_e$; the source is repeated for comparison.}
\label{fig:edit-window-grid}
\end{figure}

\paragraph{Noise-window and strength sweep.}
Under this convention $t$ increases with noise. Figure~\ref{fig:edit-window-grid} uses one puppy source, one target prompt, a fixed checkpoint and two edit coefficients. Low-noise windows tend to produce local artifacts without the requested object; $[0.75,0.90]$ gives a clear sunglasses shape. The trend is not monotone: $[0.85,0.95]$ is weaker at $c_e=100$.

For this visualization, the source prompt describes a small brown-and-white puppy on a plain white background, and the target adds realistic black sunglasses. Each SD3 cell runs to $K=120$ with $M=2$, 16 controller rollout steps, $\tau=0.00125$, $\alpha\tau^2=1$, and $c_r=c_p=0$. The figure uses the saved $K=40$ checkpoint from each cell. The rows set $c_e\in\{100,250\}$; the columns use the six noise-time windows printed in the figure.

\begin{figure}[H]
\centering
\includegraphics[width=\linewidth,trim=0bp 9bp 0bp 14bp,clip]{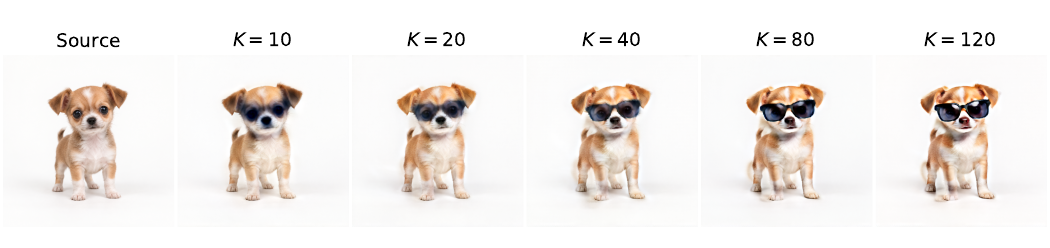}
\caption{Saved outer-loop checkpoints for the puppy sunglasses edit at $c_e=250$ and $t\in[0.75,0.90]$.}
\label{fig:edit-puppy-k}
\end{figure}

Figure~\ref{fig:edit-puppy-k} follows one of the same runs rather than restarting at each budget. The sunglasses form early and become more distinct as the controller receives further endpoint updates.

\begin{figure}[H]
\centering
\includegraphics[width=0.49\linewidth,trim=35bp 4bp 0bp 12bp,clip]{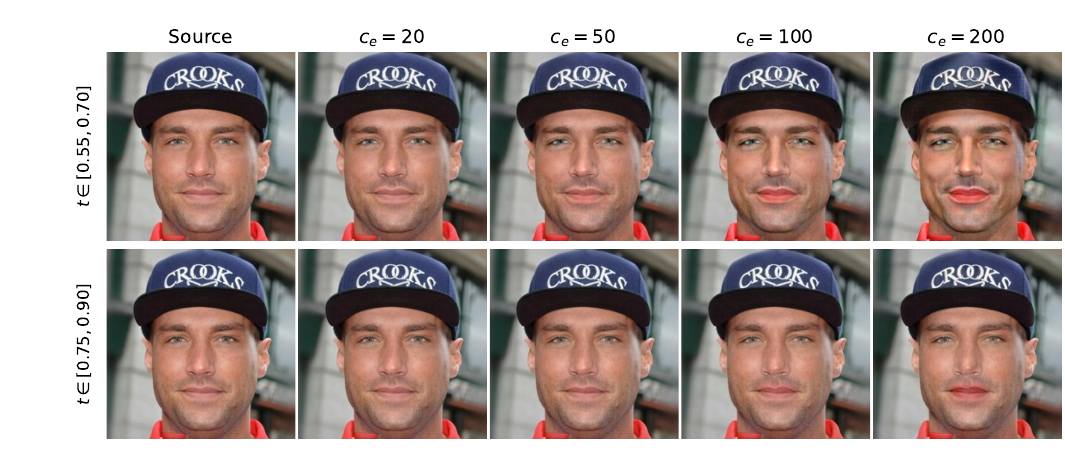}\hfill
\includegraphics[width=0.49\linewidth,trim=35bp 4bp 0bp 12bp,clip]{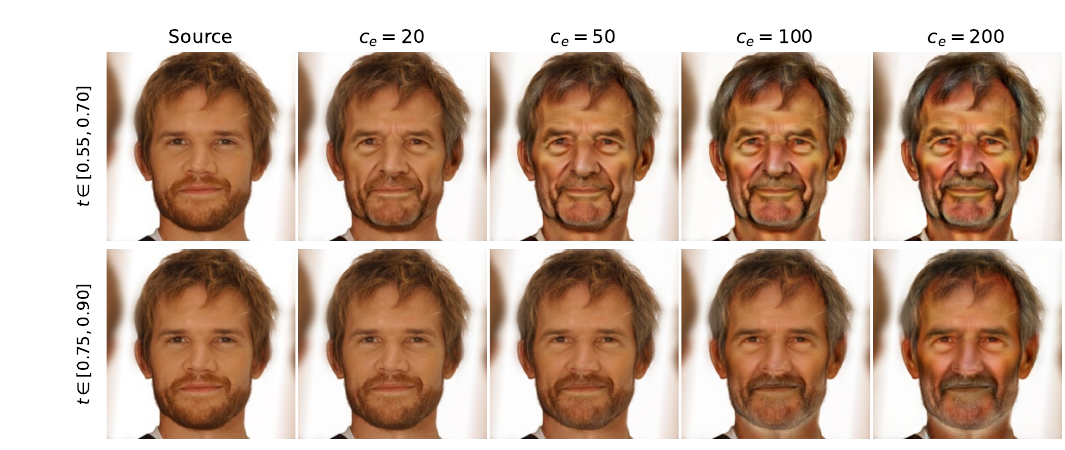}
\vspace{-5pt}
\caption{Editing Term strength and noise-time sweeps for lipstick (left) and apparent age (right). The source is repeated at left of each row.}
\label{fig:edit-face-grid}
\end{figure}

\paragraph{Additional semantic edits.}
Figure~\ref{fig:edit-face-grid} uses two visibly effective facial-attribute cases from the completed SD3 Editing Term sweep. They show saved $K=20$ checkpoints from runs with $K=40$, $M=2$, 16 controller rollout steps, $\tau=0.0125$, $\alpha\tau^2=1$, $c_r=c_p=0$, and seed zero. The four output columns vary $c_e\in\{20,50,100,200\}$; the two rows in each panel show noise-time intervals $[0.55,0.70]$ and $[0.75,0.90]$. The source and target prompts are ``A man without lipstick'' and ``A man wearing lipstick'' for the first case, and ``A young man'' and ``An old man'' for the second. These visual sweeps use the Editing Term without CLIPScore reward optimization and are separate from the four-task quantitative benchmark.

\clearpage
\section{Endpoint feedback reuse}
\label{app:feedback-reuse}
We vary the number of controller updates per feedback refresh, $M\in\{1,2,4,8,16\}$, and inspect checkpoints at $K=10,20,\ldots,100$ on four SD3 tasks. Each setting follows one continuous run on the same three source images per task, with the task-specific reward and preservation coefficients fixed. The cache holds only the latest endpoint--target pair. Thus, at fixed $K$, increasing $M$ reuses the same number of expensive endpoint targets for more small-network updates ($KM$ in total); this comparison varies supervision reuse rather than cache capacity or equal-time performance.

\begin{figure}[!ht]
\centering
\includegraphics[width=\linewidth]{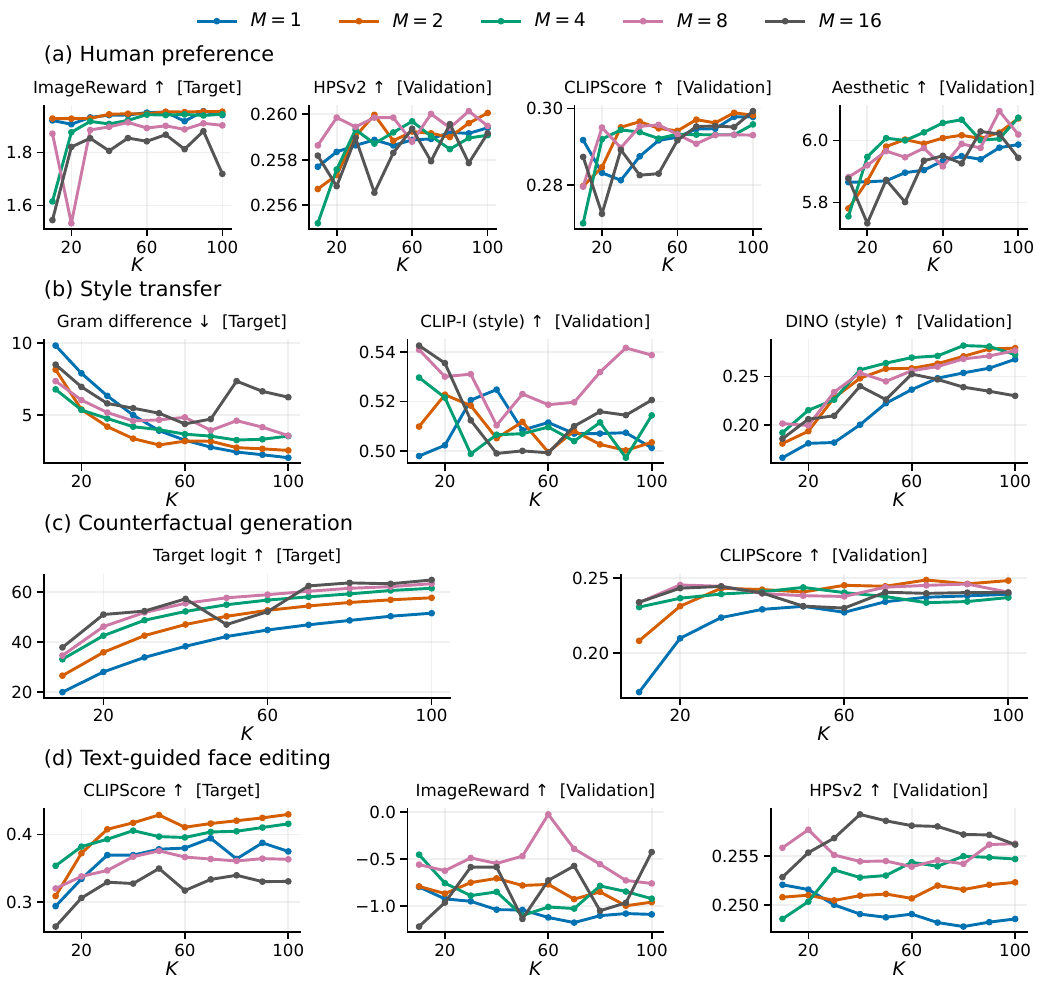}
\caption{Endpoint feedback reuse across four SD3 tasks. Curves show three-image means as the number of outer refreshes $K$ increases; colors denote the number of controller updates $M$ per refresh. The first panel in each row is the optimized target, and the remaining panels are validation metrics.}
\label{fig:feedback-reuse}
\end{figure}

Figure~\ref{fig:feedback-reuse} reports target and validation metrics. Moderate reuse improves the target at some query budgets, but increasing $M$ is not uniformly beneficial: at later checkpoints, $M=2$ gives stronger ImageReward and face-editing targets than the largest reuse settings, while style transfer favors fewer updates. Counterfactual target logits benefit from more reuse, illustrating that the preferred setting depends on the task. These curves characterize the selected examples rather than the full benchmark.

\FloatBarrier

\end{document}